\documentclass{article}

\PassOptionsToPackage{numbers, compress}{natbib}
\usepackage[final]{neurips_2025}

\usepackage[utf8]{inputenc} 
\usepackage[T1]{fontenc}    
\usepackage{hyperref}       
\usepackage{url}            
\usepackage{booktabs}       
\usepackage{amsfonts}       
\usepackage{nicefrac}       
\usepackage{microtype}      
\usepackage{xcolor}         

\usepackage{amsmath,amsfonts}
\usepackage{array}
\usepackage[justification=centering]{caption}
\usepackage[caption=false,font=normalsize,labelfont=sf,textfont=sf]{subfig}
\usepackage{textcomp}
\usepackage{stfloats}
\usepackage{url}
\usepackage{verbatim}
\usepackage{graphicx}
\usepackage{multirow}
\usepackage{amsthm}
\usepackage{amssymb}
\usepackage{mathtools}
\usepackage{xspace}
\usepackage{float}
\usepackage{bm}
\usepackage{algorithm}
\usepackage{algorithmicx}
\usepackage[noend]{algpseudocode}
\usepackage{bbding}
\usepackage{pifont}
\usepackage{wrapfig} 

\usepackage{color}
\usepackage {colortbl,array,xcolor}

\usepackage{xspace} 
\newcommand{\method}{Binary Quadratic Quantization\xspace}
\newcommand{\abbrevmethod}{BQQ\xspace}

\algnewcommand\LeftComment[2]{%
\hspace{#1\algindent}$\triangleright$ \eqparbox{COMMENT}{#2} \hfill 
}
\algnewcommand\algorithmicto{\textbf{to}}
\algnewcommand\algorithmicin{\textbf{in}}
\algnewcommand\algorithmicpara{\textbf{parallel}}
\algdef{SE}[FOR]{ForTo}{EndForTo}[2]
    {\algorithmicfor\ #1\ \algorithmicto\ #2\ \algorithmicdo}
    {\algorithmicend\ \algorithmicfor}
\algdef{SE}[FOR]{ForIn}{EndForIn}[2]
    {\algorithmicfor\ #1\ \algorithmicin\ #2\ \algorithmicdo}
    {}
\algdef{SE}[FOR]{ForToPara}{EndForToPara}[2]
    {\algorithmicfor\ #1\ \algorithmicto\ #2\ \algorithmicdo\ \algorithmicin\ \algorithmicpara}
    {\algorithmicend\ \algorithmicfor}
\algtext*{EndForTo}
\algtext*{EndForIn}
\algtext*{EndForToPara}

\usepackage{lineno}

\title{Binary Quadratic Quantization: Beyond First-Order Quantization for Real-Valued Matrix Compression}

\author{
  Kyo Kuroki \\
  Institute of Science Tokyo \\
  \texttt{kuroki.kyo@artic.iir.isct.ac.jp}
  \And
  Yasuyuki Okoshi \\
  Institute of Science Tokyo \\
  \texttt{okoshi.yasuyuki@artic.iir.isct.ac.jp}
  \And
  Thiem Van Chu \\
  Institute of Science Tokyo \\
  \texttt{thiem@artic.iir.isct.ac.jp}
  \And
  Kazushi Kawamura \\
  Waseda University \\
  \texttt{kawamura.k.au@waseda.jp}
  \And
  Masato Motomura \\
  Institute of Science Tokyo \\
  \texttt{motomura@artic.iir.isct.ac.jp}
}

\begin{document}

\maketitle

\begin{abstract}
This paper proposes a novel matrix quantization method, \method (\abbrevmethod). In contrast to conventional first-order quantization approaches---such as uniform quantization and binary coding quantization---that approximate real-valued matrices via linear combinations of binary bases, \abbrevmethod leverages the expressive power of binary quadratic expressions while maintaining an extremely compact data format.
We validate our approach with two experiments: a matrix compression benchmark and post-training quantization (PTQ) on pretrained Vision Transformer-based models.
Experimental results demonstrate that \abbrevmethod consistently achieves a superior trade-off between memory efficiency and reconstruction error than conventional methods for compressing diverse matrix data. It also delivers strong PTQ performance, even though we neither target state-of-the-art PTQ accuracy under tight memory constraints nor rely on PTQ-specific binary matrix optimization.
For example, our proposed method outperforms the state-of-the-art PTQ method by up to 2.2\% and 59.1\% on the ImageNet dataset under the calibration-based and data-free scenarios, respectively, with quantization equivalent to 2 bits.
These findings highlight the surprising effectiveness of binary quadratic expressions for efficient matrix approximation and neural network compression.

\end{abstract}

\vspace{-0.75em}
\section{Introduction}

Modern information systems increasingly demand efficiency in both computation and resource usage, driven by growing model sizes, data volumes, and deployment requirements across diverse hardware environments. 
In these systems, real-valued matrices frequently appear as weight parameters in deep neural networks (DNNs), as high-dimensional embeddings in retrieval systems, and as training datasets.
Because such matrices are central to a wide range of data processing and applications, their efficient representation and compression is crucial for reducing the costs of storage, computation, and data movement---an essential step toward deploying models on edge devices, reducing memory usage in retrieval systems, or scaling to large datasets learning.

A widely adopted strategy for this purpose is quantization, which approximates continuous-valued data with discrete levels to save memory and enable faster computation. Most existing methods rely on first-order scalar quantization approaches—such as Uniform Quantization (UQ) or Binary Coding Quantization (BCQ)~\cite{DBLP_CVPR_CNN_BCQ}—that represent real-valued matrices as linear combinations of binary bases.  While effective under moderate compression, such first-order methods often struggle to accurately reconstruct the original matrix with ultra-low-bit quantization, as the number of possible values for each element becomes extremely limited.
Beyond these scalar quantization approaches, alternative techniques such as Vector Quantization (VQ)~\cite{gersho1991_vq} and its variants (e.g., Product Quantization (PQ)~\cite{jegou2011_pq} and Lattice Vector Quantization (LVQ)~\cite{Agrell1998_lvq, GIBSON1988_lvq, Vaishampayan2001_lvq}), as well as low-rank approximations~\cite{Eckart1936_low_rank}, have also been explored. While these methods can capture correlations among dimensions more effectively, they typically rely on codebooks or factorized components that contain unquantized real values. Consequently, although the index or low-rank representation is compact, their dependence on floating-point vectors limits hardware efficiency, unlike scalar quantization methods.
Although several studies~\cite{dettmers2023qlora, guo2024lqlora, ICLR2024_QA-LoRA, saha2024_caldera} have explored combining low-rank approximation with scalar quantization, the use of extremely low-bit (e.g., binary) representations for the factorized components has not yet been fully explored.
Furthermore, to our knowledge, no prior work has applied, as in BCQ, independent scaling factors for each binary matrix in a factorized representation.
Incorporating such strategies could potentially enable even more efficient matrix approximation under extreme low-bit constraints.
Motivated by these observations, we introduce Binary Quadratic Quantization (BQQ), a framework that represents matrices using quadratic combinations of binary variables, with independent scaling factors assigned to each binary matrix.
Specifically, the target matrix is represented as a sum of binary matrix products, enabling expressive nonlinear approximations while maintaining an exceptionally compact data format.
This approach pushes the boundaries of matrix quantization by addressing the limitations of traditional methods and offering a fundamentally new perspective on matrix approximation.

In this paper, we demonstrate the effectiveness of \abbrevmethod through comprehensive evaluations: (i) measuring the trade-off between quantization error and memory usage across various matrix datasets, (ii) assessing performance when applied to post-training quantization (PTQ) of DNNs.
Beyond these applications, the generality of our framework suggests its potential for other scenarios where efficient matrix approximations are essential, such as accelerating approximate nearest neighbor (ANN)~\cite{Sunil_ANN, Malkov_HNSW}-based retrieval systems and improving the scalability of large-scale learning powered by abundant training data~\cite{zhao_dataset_quantization, zhou2023datasetquantization}.

Our main contributions are:
\vspace{-0.5em}
\begin{itemize}
\item We propose \abbrevmethod, a novel matrix quantization framework based on quadratic expressions of binary matrices, offering a new perspective on extreme matrix compression.
\item Minimizing the quantization error under the \abbrevmethod formulation naturally leads to an NP-hard optimization problem. To address this issue, we develop an efficient solution based on polynomial unconstrained binary optimization (PUBO) and convex quadratic programming.
\item We demonstrate that \abbrevmethod consistently achieves an excellent trade-off between memory usage and quantization error for compressing diverse matrix data.
\item We show that \abbrevmethod also delivers state-of-the-art (SOTA) performance in weight PTQ of Vision Transformer~\cite{dosovitskiy2021an_ViT}-based models (ViTs), even though our PTQ method is based mainly on minimizing weight reconstruction error, rather than explicitly minimizing activation error to achieve SOTA performance.

To the best of our knowledge, this is the first study to achieve practical accuracy--such as 72\% ImageNet top-1 accuracy on the DeiT-base model--using data-free PTQ for ViTs at a model size equivalent to 2-bit quantization.
\end{itemize}

\vspace{-1em}
\section{Related Works}
\vspace{-0.4em}
\paragraph{Quantization}

Quantization is a fundamental technique for reducing the precision of real-valued parameters, widely used for model compression and efficient processing. It converts continuous values into a limited set of discrete levels, with methods varying in granularity and complexity depending on hardware constraints and the acceptable accuracy-performance trade-off.
UQ is the most commonly used quantization method, which approximates a real-valued matrix $\mathbf{W} \in \mathbb{R}^{m \times n}$ as:
\small
\begin{equation}
    \mathbf{W} \approx a \sum_{i=0}^{p-1} 2^i \mathbf{B}_i + b\boldsymbol{1}, 
    \label{eq:uniform_quantization_def}
\end{equation}
\normalsize
where $\mathbf{B}_i \in \{0, 1\}^{m \times n}$, $a \in \mathbb{R}$ is the scaling factor, and $b \in \mathbb{R}$ is a bias (or zero-point). This form corresponds to $p$-bit quantization with uniform step sizes.
On the other hand, non-uniform quantization assigns quantization levels in a data-aware manner, allowing better alignment with the underlying distribution of matrices and reducing quantization error. One example is BCQ~\cite{DBLP_CVPR_CNN_BCQ}, which approximates a real-valued matrix as a sum of binary bases with individual scaling factors:
\small
\begin{equation}
    \mathbf{W} \approx \sum_{i=0}^{p-1} a_i \mathbf{B}_i,
    \label{eq:bcq_def}
\end{equation}
\normalsize
where $a_i \in {\mathbb{R}}$ and $\mathbf{B}_i \in \{0,1\}^{m \times n}$~\cite{chen2024_double_binary} or $\{-1,1\}^{m \times n}$~\cite{xu2018_BCQ, Jeon_2022_CVPR_BCQ, kwon2021posttraining_BCQ, DBLP_CVPR_CNN_BCQ, you2024_shiftaddllm, bulat2024_qbb, DBLP_ICLR2024_LUT_GEMM}. 
By introducing a bias term, the $\{0,1\}^{m \times n}$ and $\{-1,1\}^{m \times n}$ representations can be made equivalent and encompass UQ. This enables flexible quantization levels that can better capture the characteristics of $\mathbf{W}$.
Alternatively, some methods apply a nonlinear transformation before quantization. One such method is logarithmic quantization~\cite{MiyashitaLM16_CNN_log_quantization}, which approximates the logarithmic scale of the original weights:
$
    \text{log}_2|\mathbf{W}| \approx \sum_{i=0}^{p-1} 2^i \mathbf{B}_i + b\boldsymbol{1},
$
where $\mathbf{B}_i \in \{0,1\}^{m \times n}$ and $b \in {\mathbb{R}}$. This technique is particularly effective when the distribution of matrix values is highly skewed. It also offers a hardware-friendly implementation of matrix multiplication, as the powers-of-two representation allows the operation to be performed using efficient bit-shift operations.
 However, because the sign information is lost in the logarithmic transformation, an additional bit is required to retain the original sign of each element.

\vspace{-0.4em}
\paragraph{Matrix Factorization}

Matrix factorization expresses a matrix $\mathbf{W}^{m \times n}$ exactly or approximately as a product $\mathbf{W} \approx \mathbf{Y} \mathbf{Z}$, and is a fundamental tool in signal processing, machine learning, and data analysis. The target matrix $\mathbf{W}^{m \times n}$, and the factor matrices $\mathbf{Y}^{m \times l}$ and $\mathbf{Z}^{l \times n}$ are subject to different constraints depending on the specific method. For example:
\vspace{-0.5em}
\small
\begin{itemize}
    \item \textbf{Singular Value Decomposition (SVD)}~\cite{Golub1970_SVD}: $\mathbf{W} \in \mathbb{R}^{m \times n}$, $\mathbf{Y} \in \mathbb{R}^{m \times l}$ , $\mathbf{Z} \in \mathbb{R}^{l \times n}$ ,\\
    Note that singular value diagonal matrix and orthogonal matrix are described as one.
    \item \textbf{Non-negative Matrix Factorization (NMF)}~\cite{NIPS2000_NMF}: $\mathbf{W} \in \mathbb{R}_{\geq 0}^{m \times n}$, $\mathbf{Y} \in \mathbb{R}_{\geq 0}^{m \times l}, \mathbf{Z} \in \mathbb{R}_{\geq 0}^{l \times n}$
    \item \textbf{Real/Binary Matrix Factorization (RBMF)}~\cite{NIPS2013_RBMF}: $\mathbf{W} \in \mathbb{R}^{m \times n}$, $\mathbf{Y} \in \mathbb{R}^{m \times l}$, 
    $\mathbf{Z} \in \{0,1\}^{l \times n}$
    \item \textbf{Non-negative/Binary Matrix Factorization (NBMF)}~\cite{OMalley2018_NBMF}: $\mathbf{W} \in \mathbb{R}_{\geq 0}^{m \times n}$, $\mathbf{Y} \in \mathbb{R}^{m \times l}_{\geq 0}$, $\mathbf{Z} \in \{0,1\}^{l \times n}$
    \item \textbf{Binary Matrix Factorization (BMF)}~\cite{Zhang2007_BMF}: $\mathbf{W} \in \{0,1\}^{m \times n}$, $\mathbf{Y} \in \{0,1\}^{m \times l}, \mathbf{Z} \in \{0,1\}^{l \times n}$
    \item \textbf{Boolean Matrix Factorization (BoolMF)}~\cite{miettinen2006_boolmf, article_BoolMF}: $\mathbf{W} \in \{0,1\}^{m \times n}$, $\mathbf{Y}\in \{0,1\}^{m \times l}, \mathbf{Z}^{l \times n} \in \{0,1\}$, with Boolean product: $\mathbf{W}_{ij} \approx \bigvee_{k=1}^l (\mathbf{Y}_{ik} \land \mathbf{Z}_{kj})$
\end{itemize}
\normalsize
Such matrix factorization techniques are widely used not only in data analysis but also for matrix compression through low-rank approximation (i.e., $l \ll \min(m, n)$)~\cite{Eckart1936_low_rank}.
Building on this idea, Low-Rank Adaptation (LoRA)~\cite{hu2022lora} enables efficient fine-tuning of large pre-trained models by restricting weight updates to a low-rank subspace. More recently,~\cite{dettmers2023qlora, ICLR2024_QA-LoRA, guo2024lqlora} combines LoRA with quantization, further reducing memory usage while preserving model quality, and has become a widely adopted approach for resource-efficient fine-tuning.

\vspace{-0.5em}
\section{Preliminaries}

\vspace{-0.5em}
\paragraph{Polynomial Unconstrained Binary Optimization (PUBO)}

Polynomial Unconstrained Binary Optimization (PUBO)~\cite{glover_pubo_part1, glover_pubo_part2} is a class of combinatorial optimization problems defined as the minimization of a multivariate polynomial over binary variables. Formally, it can be expressed as:
\small
\begin{equation}
    \label{eq:PUBO_function}
    L(\boldsymbol{s})
    = \sum_{i_1} J_{i_1}^{(1)} s_{i_1}
    + \sum_{i_1<i_2} J_{i_1i_2}^{(2)} s_{i_1}s_{i_2}
    + \cdots
    + \sum_{i_1<i_2<\cdots<i_k} J_{i_1i_2\cdots i_k}^{(k)} \prod_{j=1}^{k} s_{i_j},
\end{equation}
\normalsize
where $\boldsymbol{s} \in \{0,1\}^N$ is a binary vector and $J^{(k)}$ denotes the $k$-th order interaction coefficients. In the special case where the degree $k=2$, the problem reduces to the well-known Quadratic Unconstrained Binary Optimization (QUBO) formulation~\cite{qubo_tutorial, qubo_survey}.
QUBO is equivalent to minimizing the energy function of the Ising model~\cite{Ising_1925} and has been widely studied in various fields, including physics, computer science, and artificial intelligence. In general, solving PUBO, including QUBO, problems is NP-hard, and thus, a broad range of heuristics~\cite{SB_pubo, Goto_19_SB, Goto_21_bSB, Chermoshentsev2021_simCIM_pubo, kuroki2025_AMFD, ma2024NIPS_HD}, optical computing~\cite{Inagaki_16_CIM, Honjo_21_CIM_100000spin} and quantum computing~\cite{Kadowaki_98_QA, Johnson_11_QA_DWave} have been proposed to tackle them efficiently.

\vspace{-0.5em}
\paragraph{Annealed Mean Field Descent}

\begin{algorithm}[b]
\caption{One Iteration of AMFD~\cite{kuroki2025_AMFD}} \label{Alg:AMFD}

\begin{algorithmic}[1]
\Require $L$, $\bm{x}_\text{old}$,  $\bm{x}_\text{cur}$, $T$, $\Delta T$, $\eta$,  $\zeta$
\Ensure $\boldsymbol{x}_\text{cur}, \boldsymbol{x}_\text{new}$, $T$
    \State $\boldsymbol{x}_\text{fwd} \gets \boldsymbol{x}_\text{cur} + \zeta\left(\boldsymbol{x}_\text{cur} - \boldsymbol{x}_\text{old}\right)$
    \Comment{Forward point}
        \State $\boldsymbol{{\Phi}} \gets \nabla L(\boldsymbol{x}_\text{fwd})$
            \Comment{Gradient of the first term in Eq.~\eqref{eq:KL_pubo}}, scaled by $T$
        \State $\boldsymbol{F} \gets T\left(\boldsymbol{x}_\text{cur}-\boldsymbol{0.5}\right)$ \Comment{Gradient of the other terms in Eq.~\eqref{eq:KL_pubo} (approx. expr.) scaled by $T$}
        \State $\boldsymbol{x}_\text{new} \gets \texttt{clip}\left(2{\boldsymbol{x}}_\text{cur} - \boldsymbol{x}_\text{old} - \eta \left({\boldsymbol{F}} + \boldsymbol{\Phi}\right), 0, 1\right)$ 
        \Comment{Descent with acceleration and constraints}
    \State $T \gets T-\Delta T$ \Comment{Annealing}
\end{algorithmic}
\vspace{-0.2em}
\end{algorithm}
\normalsize

One of the recent promising approaches for solving QUBO problems is Annealed Mean Field Descent (AMFD)~\cite{kuroki2025_AMFD}, which is based on Mean Field Annealing~\cite{NIPS1988_ec5decca_MFA}. 
It aims to find minimum solutions by gradually annealing the temperature while optimizing a mean-field approximation to the canonical distribution:
$
    P_{\text{C}}(\boldsymbol{s}) = \frac{1}{Z} \exp\left(-\frac{L(\boldsymbol{s})}{T}\right), \ Z = \sum_{\boldsymbol{s}} \exp\left(-\frac{L(\boldsymbol{s})}{T}\right).
$
Since computing the distribution is generally intractable due to the exponential number of configurations, it is approximated by an independent distribution for each variable (mean-field approximation):
$
    P_{\text{MF}}(\boldsymbol{s}) = \prod_{i=1}^{N} p_i(s_i),
$
where $p_i(s_i)$ denotes the probability of taking value $s_i$.
AMFD minimizes the Kullback–Leibler (KL) divergence between $P_\text{MF}(\boldsymbol{s})$ and $P_\text{c}(\boldsymbol{s})$ by gradient descent-based updates. 
At low temperatures, the canonical distribution concentrates on minimum states, thereby allowing the extraction of approximate minimum solutions.
While AMFD derived an explicit form of the KL divergence for QUBO problems, this work extends the framework to general PUBO problems. We show that the KL divergence between $P_{\text{MF}}(\boldsymbol{s})$ and $P_{\text{C}}(\boldsymbol{s})$ can be written as:
\begin{equation}\label{eq:KL_pubo}
    D_{\text{KL}}\left(P_{\text{MF}}(\boldsymbol{s}) \,\|\, P_{\text{C}}(\boldsymbol{s}) \right) 
    =\frac{L(\boldsymbol{x})}{T} + \ln Z + \sum_{i=1}^{N} \left[ (1-x_{i}) \ln (1-x_i)+x_i\ln x_i\right]
    ,
\end{equation}
where $x_i = \sum_{s_{i}=0}^{1} s_{i}p_{i}(s_{i})=p_{i}(1)$ is the expectation of $s_i$ under the mean-field distribution. 
Please refer to the App.~\ref{app:appendix_kl_pubo_proof} for a detailed derivation of this formulation.
Note that $x_i$ is a real-valued variable ranging from 0 to 1, rather than a binary variable. Therefore, the KL divergence is differentiable with respect to $x_i$.
One iteration of the AMFD algorithm is illustrated in Alg.~\ref{Alg:AMFD}.
Note that the term $\left[ (1 - x_i) \ln (1 - x_i) + x_i \ln x_i \right]$ in Eq.~\eqref{eq:KL_pubo} is approximated by a second-order Taylor expansion around $x_i = 0.5$
to prevent numerical overflow when $x_i$ is close to 0 or 1.
This paper applies the extended AMFD to optimize quantized representations under the general PUBO setting.

\section{Proposed Method}
\label{sec:method}

\subsection{\method (\abbrevmethod)}\label{subsec:binary_quadratic_quantization}
A primary limitation of conventional first-order quantization methods like UQ and BCQ is the limited number of distinct values each element can take. For example, 1-bit quantization allows only two levels (e.g., $\{-1, +1\}$ or $\{0, 1\}$), and 2-bit quantization increases this to just four. Such coarse granularity restricts representational flexibility, especially under aggressive compression.
While such methods are limited in expressiveness, binary matrix multiplication can yield outputs with a wider value range, enabling multi-bit representations even though each binary matrix individually encodes only minimal information. This property suggests a previously underexplored potential for approximating real-valued matrices through compositions of binary matrices, offering a new perspective beyond traditional quantization methods.
Nonetheless, existing matrix decomposition approaches operate within fixed numerical domains---either real-to-real (e.g., SVD, NMF) or binary-to-binary (e.g., BMF, BoolMF). Hybrid methods like RBMF and NBMF bridge these domains but stop short of fully binary decompositions of real-valued matrices.

Motivated by this gap, we explore a novel quantization scheme that, unlike BCQ which uses linear combinations of binary matrices, is based on linear combinations of binary matrix products:
\small
\begin{equation}\label{eq:before_BQQ}
\boldsymbol{W} \approx \sum_{i=0}^{p-1}(\alpha_i \boldsymbol{Y}_i + \beta_i\boldsymbol{1}_Y)(\gamma_i \boldsymbol{Z}_i + \delta_i\boldsymbol{1}_Z),
\end{equation}
\normalsize
where $\boldsymbol{W} \in \mathbb{R}^{m \times n}$, $\boldsymbol{Y}_i \in \{0, 1\}^{m \times l}$ and $\boldsymbol{Z}_i \in \{0, 1\}^{l \times n}$ are binary matrices, while $\alpha_i,\gamma_i \in \mathbb{R}$ are scaling factors, and $\beta_i, \delta_i \in \mathbb{R}$ are bias terms.
Also, $\mathbf{1}_Y, \mathbf{1}_Z, \mathbf{1}$ denote all-ones matrices with the same shape as $\mathbf{Y}, \mathbf{Z}, \mathbf{W}$, respectively.
Notably, this formulation subsumes BCQ as a special case. Specifically, when $l = \max(m, n)$, setting $\alpha_i\boldsymbol{Y}_i+\beta_i\boldsymbol{1}_Y$ as the identity matrix (if $m \geq n$), or setting $\gamma_i \boldsymbol{Z}_i + \delta_i\boldsymbol{1}_Z$ as the identity matrix (if $m \leq n$), recovers the standard BCQ structure.
We now turn our attention to the generalized form of Eq.~\eqref{eq:before_BQQ}, referred to as \method (\abbrevmethod):
\small
\begin{equation}\label{eq:qq_approx}
    \boldsymbol{W} \approx \sum_{i=0}^{p-1}\left(r_i\boldsymbol{Y}_i\boldsymbol{Z}_i + s_i\boldsymbol{Y}_i\mathbf{1}_Z + t_i\mathbf{1}_Y\boldsymbol{Z}_i\right) + u\mathbf{1}, 
\end{equation}
\normalsize
where $r_i, s_i, t_i,u \in \mathbb{R}$ are scalar coefficients.
Note that the all-one matrices are used only for notational convenience and do not need to be stored explicitly; only the binary matrices and scalar coefficients must be preserved. Also, the intermediate dimension $l$ can be arbitrarily set, allowing the number of binary elements to be adjusted independently of the original matrix size.


\vspace{-0.5ex}
\subsection{Mixed Integer Programming for \abbrevmethod}
\vspace{-0.5ex}
\begin{algorithm}[b]
\caption{Subproblem Solving via AMFD}
\label{alg:subproblem}
\begin{algorithmic}[1]
\Require Input matrix $\mathbf{W} \in \mathbb{R}^{m \times n}$, initial temperature $T_{\text{init}}$, final temperature $T_{\text{fin}}$, steps $N_{\text{step}}$, learning rate $\eta$, accelerating rate $\zeta$, intermediate dimension $l$
\Ensure Binary matrices ${\mathbf{Y}} \in \{0,1\}^{m \times l}$, ${\mathbf{Z}} \in \{0,1\}^{l \times n}$, scaling factors $r_i, s_i, t_i, u_i \in \mathbb{R}$

\State Let $\mathbf{1} \in \{1\}^{m \times n}$, $\mathbf{1}_Y \in \{1\}^{m \times l}$, and $\mathbf{1}_Z \in \{1\}^{l \times n}$
\State Sample $\hat{\mathbf{Y}}_{\text{old}}, \hat{\mathbf{Z}}_{\text{old}} \sim \mathcal{U}(0,1)$ \Comment{Initial expectation values}
\State $\hat{\mathbf{Y}} \gets \hat{\mathbf{Y}}_{\text{old}} - \eta (\hat{\mathbf{Y}}_{\text{old}} - \mathbf{0.5})$,
\ $\hat{\mathbf{Z}} \gets \hat{\mathbf{Z}}_{\text{old}} - \eta (\hat{\mathbf{Z}}_{\text{old}} - \mathbf{0.5})$
 \State $\mathbf{W} \gets \mathbf{W} / \left(\max(\mathbf{W}) - \min(\mathbf{W})\right)$\Comment{Normalization}
\State $\Delta T \gets (T_{\text{init}} - T_{\text{fin}}) / (N_{\text{step}} - 1)$
\State $T \gets T_{\text{init}}$
\State $[r_i, s_i, t_i, u_i] \gets \texttt{SFO}(\hat{\mathbf{Y}}, \hat{\mathbf{Z}}, \mathbf{W})$ 
\Comment{\texttt{SFO}: scaling factors optimization using Eq.~\eqref{eq:sfo}}
\For{$t = 1$ to $N_{\text{step}}$}
    \State $[\hat{\mathbf{Y}}_{\text{old}}, \hat{\mathbf{Z}}_{\text{old}}], [\hat{\mathbf{Y}}, \hat{\mathbf{Z}}], T$
     $\gets \texttt{AMFD}\left(L^{(i)}_{\text{pubo}}(\mathbf{W}, [r_i, s_i, t_i, u_i]), [\hat{\mathbf{Y}}_{\text{old}}, \hat{\mathbf{Z}}_{\text{old}}], [\hat{\mathbf{Y}}, \hat{\mathbf{Z}}], T, \Delta T, \eta, \zeta\right)$
    \Statex\Comment{\texttt{AMFD}: AMFD iteration using Alg.~\ref{Alg:AMFD}}

    \State $[r_i, s_i, t_i, u_i] \gets \texttt{SFO}(\hat{\mathbf{Y}}, \hat{\mathbf{Z}}, \mathbf{W})$
\EndFor

\State ${\mathbf{Y}} \gets \texttt{step}(\hat{\mathbf{Y}} - \mathbf{0.5})$, \ ${\mathbf{Z}} \gets \texttt{step}(\hat{\mathbf{Z}} - \mathbf{0.5})$ \Comment{Binarization to the higher probability}
\State $[r_i, s_i, t_i, u_i] \gets \left(\max(\mathbf{W}) - \min(\mathbf{W})\right) \cdot \texttt{SFO}({\mathbf{Y}}, {\mathbf{Z}}, {\mathbf{W}})$
\end{algorithmic}
\end{algorithm}
\normalsize

To realize \abbrevmethod formulation (Eq.~\eqref{eq:qq_approx}), we consider minimizing the squared error between the original real-valued matrix and its approximation. The objective function is given by:
\small
\begin{equation}\label{eq:qq_objective}
    L_\text{\abbrevmethod} = \left\| \boldsymbol{W} - \left[ \sum_{i=0}^{p-1} \left( r_i \boldsymbol{Y}_i \boldsymbol{Z}_i + s_i \boldsymbol{Y}_i \mathbf{1}_Z + t_i \mathbf{1}_Y \boldsymbol{Z}_i \right) + u \mathbf{1} \right] \right\|_2^2,
\end{equation}
\normalsize
where the goal is to optimize $3p+1$ real-valued coefficients and the elements of $2p$ binary matrices. This is a mixed-integer optimization problem and is NP-hard, making analytical solutions intractable. 

To address this, we adopt the following strategy:
\begin{enumerate}
    \item We apply greedy optimization independently to each index $i$ in Eq.~\eqref{eq:qq_objective} to mitigate the increasing complexity from a growing number of binary variables.
    \item We decouple the optimization of real-valued and binary variables, and alternate between convex quadratic optimization and PUBO.
\end{enumerate}

To perform greedy optimization for each index $i$ in Eq.~\eqref{eq:qq_objective}, we first define the residual matrix $\boldsymbol{W}_\text{res}^{(i)}$ as the difference between the original matrix $\boldsymbol{W}$ and the partial reconstruction using variables up to index $i-1$:
$
    \boldsymbol{W}_\text{res}^{(i)} = \boldsymbol{W} - \left[ \sum_{j=0}^{i-1} \left( r_j \boldsymbol{Y}_j \boldsymbol{Z}_j + s_j \boldsymbol{Y}_j \mathbf{1}_Z + t_j \mathbf{1}_Y \boldsymbol{Z}_j  + u_j \mathbf{1} \right) \right].
$
Notably, as an exception, we set $\boldsymbol{W}_\text{res}^{(0)} = \boldsymbol{W}$. Then, the $i$-th subproblem can be formulated as the minimization of the following objective:
\small
\begin{equation}\label{eq:qq_sub}
    L_\text{sub}^{(i)} = \left\| \boldsymbol{W}_\text{res}^{(i)} -  \left( r_i \boldsymbol{Y}_i \boldsymbol{Z}_i + s_i \boldsymbol{Y}_i \mathbf{1}_Z + t_i \mathbf{1}_Y \boldsymbol{Z}_i  + u_i \mathbf{1} \right) \right\|_2^2.
\end{equation}
\normalsize

Next, to minimize the objective function (Eq.~\eqref{eq:qq_sub}), we adopt an alternating optimization approach that separates continuous and binary variables. When the continuous coefficients are fixed, the problem reduces to optimizing over binary variables only. In this case, the objective can be reformulated as a PUBO by reducing powers of binary variables using $y^2 = y$, making each such term linear. 
The resulting objective takes the following form:
\small
\begin{equation}\label{eq:qq_pubo}
\begin{aligned}
    L_\text{pubo}^{(i)} =\ &L_\text{sub}^{(i)} + r_i^2 \sum\left[\boldsymbol{Y}_i\boldsymbol{Z}_i - (\boldsymbol{Y}_i \odot \boldsymbol{Y}_i)(\boldsymbol{Z}_i \odot \boldsymbol{Z}_i)\right] \\
    &+ s_i^2n \sum\left[\boldsymbol{Y}_i - (\boldsymbol{Y}_i \odot \boldsymbol{Y}_i)\right] 
    + t_i^2m \sum\left[\boldsymbol{Z}_i - (\boldsymbol{Z}_i \odot \boldsymbol{Z}_i)\right] \\
    &+ 2r_is_i \sum\left[\boldsymbol{Y}_i\boldsymbol{Z}_i - (\boldsymbol{Y}_i \odot \boldsymbol{Y}_i)\boldsymbol{Z}_i\right]
    + 2r_it_i \sum\left[\boldsymbol{Y}_i\boldsymbol{Z}_i - \boldsymbol{Y}_i(\boldsymbol{Z}_i \odot \boldsymbol{Z}_i)\right],
\end{aligned}
\end{equation}
\normalsize
where $\sum$ denotes the elementwise sum over all entries of the corresponding matrix.

On the other hand, fixing the binary matrices, the continuous coefficients $r_i, s_i, t_i, u_i$ can be optimized in closed form via the convexity of the $\ell_2$ norm.
\small
\begin{equation}\label{eq:sfo}
    [r_i, s_i, t_i, u_i] = -[v_{r_i}, v_{s_i}, v_{t_i}, v_{u_i}]\boldsymbol{H}_{L_{\text{pubo}}}^{-1},
\end{equation}
\normalsize
where $\boldsymbol{H}_{L_{\text{pubo}}} \in \mathbb{R}^{4\times 4}$ and $\boldsymbol{v} \in \mathbb{R}^4$ are the Hessian matrix and the first-order coefficients of Eq.~\eqref{eq:qq_pubo} with respect to $r_i, s_i, t_i, u_i$, respectively.
By incorporating the optimization of the scaling factors into a single iteration of the AMFD algorithm for PUBO, we aim to solve the subproblem. The complete procedure is presented in Alg.~\ref{alg:subproblem}.  
Using the solution of this subproblem, we then approximate the original real-valued matrix in a greedy manner. This overall approach is described in Alg.~\ref{alg:greedy_qq}.

\small
\begin{algorithm}[t]
\caption{Greedy \method}\label{alg:greedy_qq}
\begin{algorithmic}[1]
\Require Matrix $\boldsymbol{W}$, learning rate $\eta$, accelerating rate $\zeta$, initial temperature $T_{\text{init}}$, final temperature $T_{\text{fin}}$, steps $N_{\text{step}}$, intermediate dimension $l$, binary matrix stacks $p$
\Ensure Binary matrices $[\boldsymbol{Y}_0, \boldsymbol{Y}_1,...,\boldsymbol{Y}_{p-1}]$, $[\boldsymbol{Z}_0, \boldsymbol{Z}_1,...,\boldsymbol{Z}_{p-1}]$, scaling factors $\boldsymbol{r}, \boldsymbol{s},\boldsymbol{t} \in \mathbb{R}^p,u\in \mathbb{R}$
\State $\boldsymbol{W}_{\text{res}} \gets \boldsymbol{W}$
\For{$i = 0$ to $(p - 1)$}
    \State $ \boldsymbol{Y}_i,  \boldsymbol{Z}_i, r_i, s_i, t_i, u_i\gets \texttt{SS}(\boldsymbol{W}_\text{res}, T_\text{init}, T_\text{fin}, N_\text{step}, \eta, \zeta, l)$ \Comment{SS: subproblem solving via Alg.~\ref{alg:subproblem}}
    \State ${\boldsymbol{W}}_\text{res} \gets {\boldsymbol{W}}_\text{res}-\left( r_i\boldsymbol{Y}_i \boldsymbol{Z}_i + s_i \boldsymbol{Y}_i \boldsymbol{1}_Z + t_i \boldsymbol{1}_Y \boldsymbol{Z}_i + u_i\boldsymbol{1}\right)$
\EndFor
\State $u \gets \sum_{i=0}^{p-1}u_i$
\end{algorithmic}
\end{algorithm}
\normalsize

\vspace{-0.5ex}
\subsection{Post-Training Quantization via \abbrevmethod} \label{subsec:ptq_method}
\vspace{-0.5ex}

\begin{figure}[t]
    \vspace{-0.75em} 
    \centering
    \includegraphics[width=0.95\linewidth]{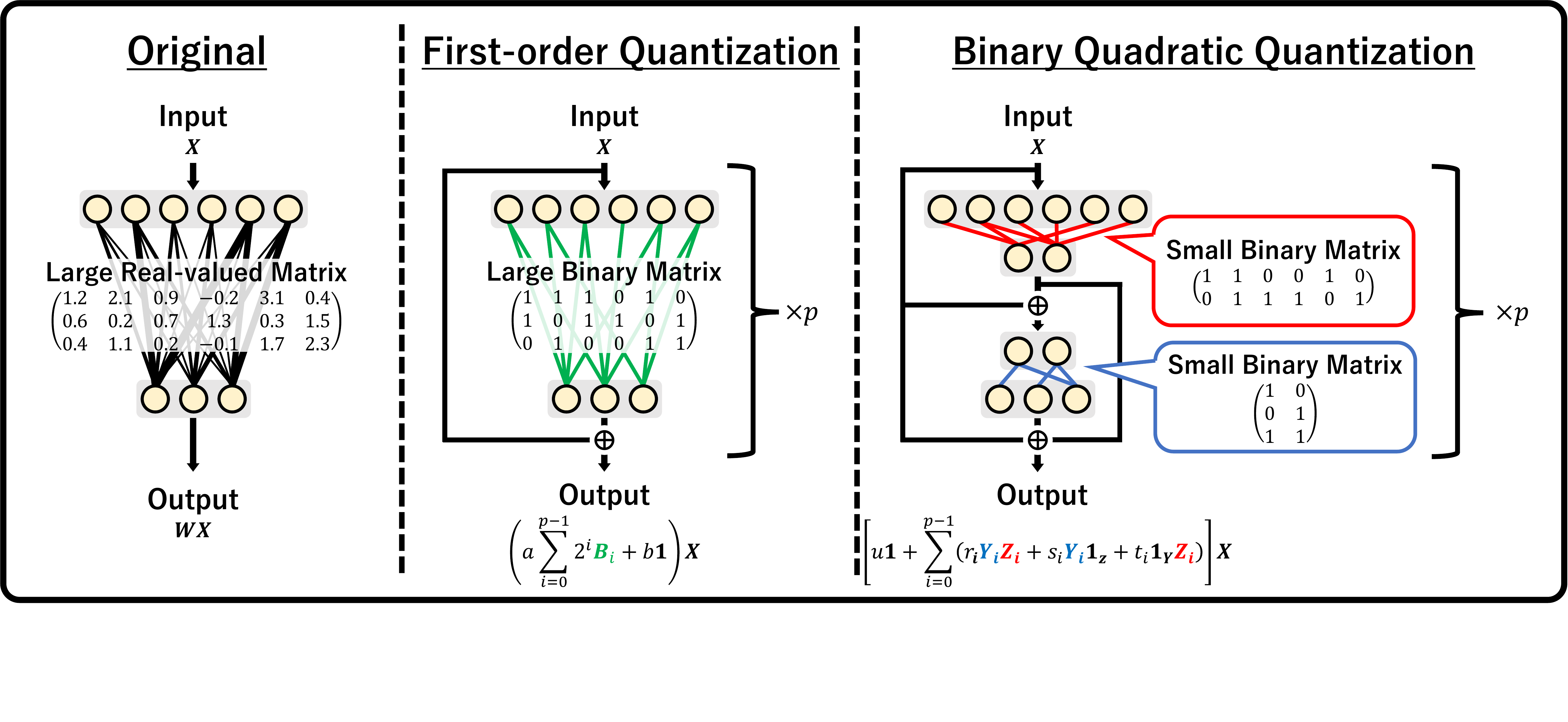}
    \vspace{-2.75em} 
    \caption{Comparison between \abbrevmethod and first-order quantization in a DNN layer.}
    \label{fig:nn_quantization}
    \vspace{-1.5em} 
\end{figure}

This subsection presents a model compression technique for deep neural networks (DNNs) based on \abbrevmethod. In particular, we focus on ViT-based models, which have recently achieved remarkable success in image processing but still struggle with ultra-low-bit quantization. Quantization methods for ViTs, as well as for general DNNs, can be categorized into two types: Quantization-Aware Training (QAT)~\cite{OFQ_ViT, Q-ViT, Xu2022TerViT}, which integrates quantization into the training process using labeled data, and Post-Training Quantization (PTQ)~\cite{ijcai2022p164_FQ-ViT, Li_2023_ICCV_RepQ-ViT, zhang2024_COMQ, ICML_ERQ, TPAMI_ERQ}, which applies quantization to a pretrained model using either unlabeled or limited data. Especially, in situations where training data is unavailable due to privacy constraints or data access limitations, the need to address these challenges has driven interest in data-free quantization techniques~\cite{PSAQ-ViT1, PSAQ_ViT2}. Our work focuses on PTQ under two different scenarios.
In the data-free setting, we perform only data-free weight quantization.
In the calibration-based setting, we first apply data-free weight quantization, followed by correction of bias and normalization parameters using a small amount of unlabeled calibration data.
Previous studies have explored weight quantization to reduce model size and inference costs, as well as quantization of both weights and activations to further reduce inference costs. In this study, we focus exclusively on weight quantization to clarify the standalone effectiveness of \abbrevmethod.
Note that, unlike standard first-order quantization, \abbrevmethod approximates the original matrix using a combination of binary matrices with altered shapes, as illustrated in Fig.~\ref{fig:nn_quantization}.

\vspace{-0.5em}
\paragraph{Data-Free Quantization}
\begin{wrapfigure}{r}{0.4\textwidth}
\vspace{-3em}
  \centering
  \includegraphics[width=0.38\textwidth]{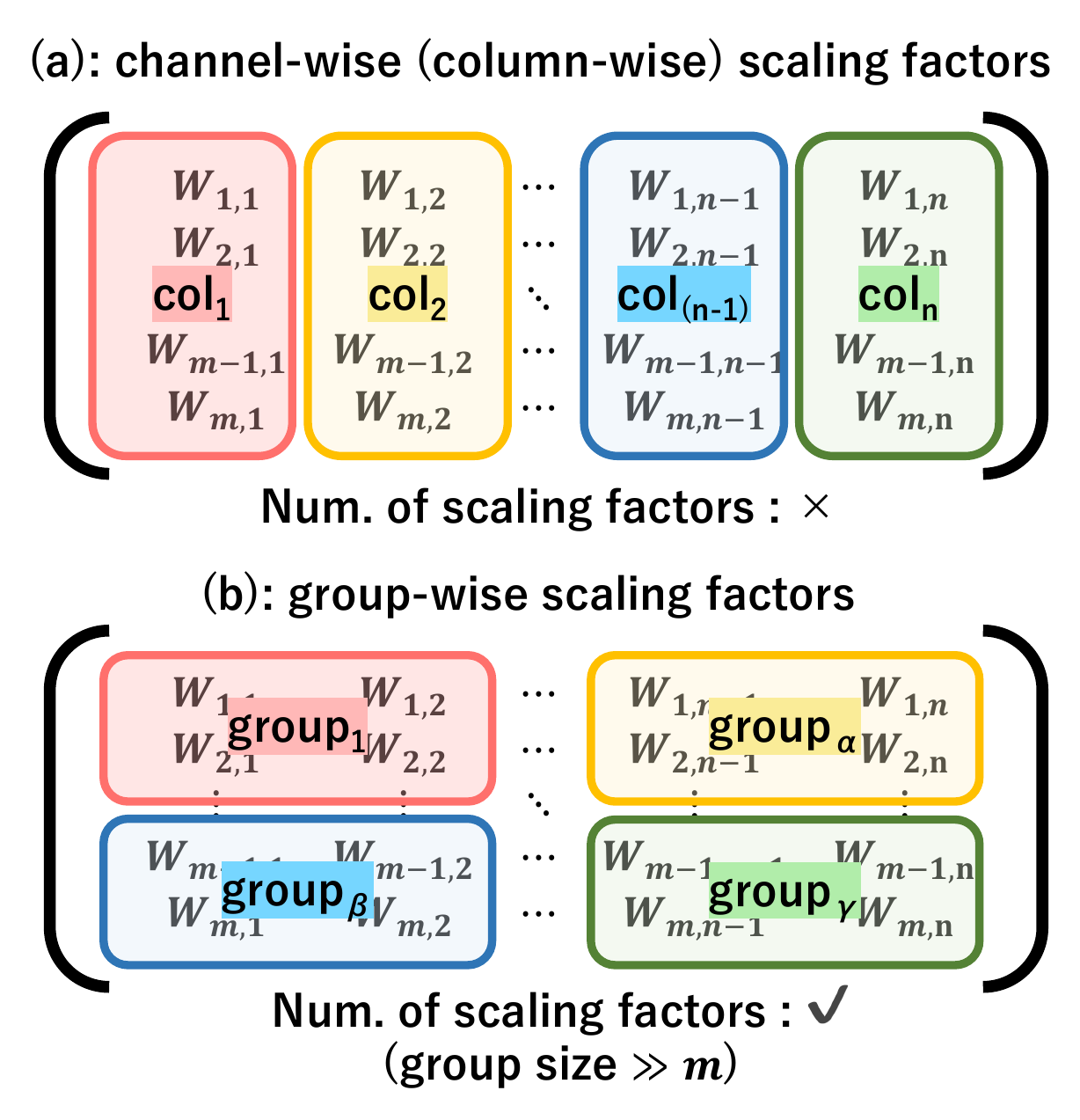}
  \vspace{-0.25em}
  \caption{Weight scaling methods.}
  \label{fig:scaling_factors}
  \vspace{-2em}
\end{wrapfigure}
First, we apply a data-free quantization approach, directly quantizing the weights. Specifically, we formulate weight quantization as an optimization problem that minimizes the squared reconstruction error between a pretrained weight matrix $\mathbf{W}$ and its quantized counterpart via \abbrevmethod, as shown in Eq.~\eqref{eq:qq_objective}.
While prior methods often use channel-wise (i.e., column-wise) scaling factors to maintain accuracy, they result in a large number of scaling parameters. Instead, we adopt a group-wise quantization strategy~\cite{you2024_shiftaddllm}, in which each weight matrix is divided into smaller submatrices, and \abbrevmethod is applied independently to each with its own set of scaling factors (see Fig.\ref{fig:scaling_factors}). This can reduce the number of scaling parameters, thereby shrinking the model size.

\vspace{-0.5em}

\paragraph{Correction of Bias and Normalization Parameters}

After quantizing all weight matrices, we optionally apply a lightweight correction step using a small set of unlabeled calibration inputs. Similar to~\cite{bulat2024_qbb}, we refine only the bias and layer normalization parameters--while keeping all other parameters fixed--by minimizing the mean squared error between the output logits of the original $f_\text{org}$ and quantized models $f_\text{q}$, as a form of knowledge distillation:

$
    \min_{\boldsymbol{\theta}} \left\| f_{\text{org}}(\boldsymbol{\theta}_\text{org}) - f_{\text{q}}(\boldsymbol{\theta}) \right\|_2^2/ \left|f_{\text{org}}(\boldsymbol{\theta}_\text{org}) \right|,
$
where $\boldsymbol{\theta}$ denotes the bias and normalization parameters. This correction step compensates for quantization-induced errors and helps recover lost accuracy without requiring full fine-tuning or access to labeled data.

\vspace{-0.25em}
\section{Evaluation}\label{sec:evaluation}
\vspace{-0.5em}
\paragraph{Implementation Details}\label{subsec:implemantation_detail}

As described in Eq.~\eqref{eq:qq_approx}, \abbrevmethod decomposes a real-valued matrix of size $m \times n$ into binary matrices $\mathbf{Y}_i \in \{0,1\}^{m \times l}$ and $\mathbf{Z}_i \in \{0,1\}^{l \times n}$. To ensure a fair comparison with baseline methods like UQ and BCQ, we fix the intermediate dimension $l = \mathrm{round}(mn / (m + n))$ for all binary matrices. This ensures the total number of binary parameters matches that of UQ and BCQ, making $p$ in Eq.~\eqref{eq:qq_approx} the pseudo bit width.
Another way to match the number of binary parameters is to adjust the ratio between the intermediate dimension $l$ and the number of stacks $p$ in Eq.~\eqref{eq:qq_approx}; however, this paper adopts the approach described above.
Unless otherwise noted, the hyperparameters used in Alg.~\ref{alg:greedy_qq} are set to the following values throughout all experiments: $T_{\text{init}} = 0.2$, $T_{\text{fin}} = 0.005$, $\eta = 0.06$, $\zeta = 4$, and $N_{\text{step}} = 50{,}000$.
Also, the scaling factor and the bias for UQ are optimized via grid search to minimize the mean squared error (MSE), as described in App.~\ref{subsec:appendix_uq_alg}. For BCQ, we implement the method based on~\cite{xu2018_BCQ}, referring to parts of the open-source code provided in~\cite{you2024_shiftaddllm}.

\vspace{-0.5em}
\paragraph{Matrix Data Compression}\label{subsec:matrix_data_compression}
\begin{figure}[t]
    \vspace{-1em} 
    \centering
    \includegraphics[width=\linewidth]{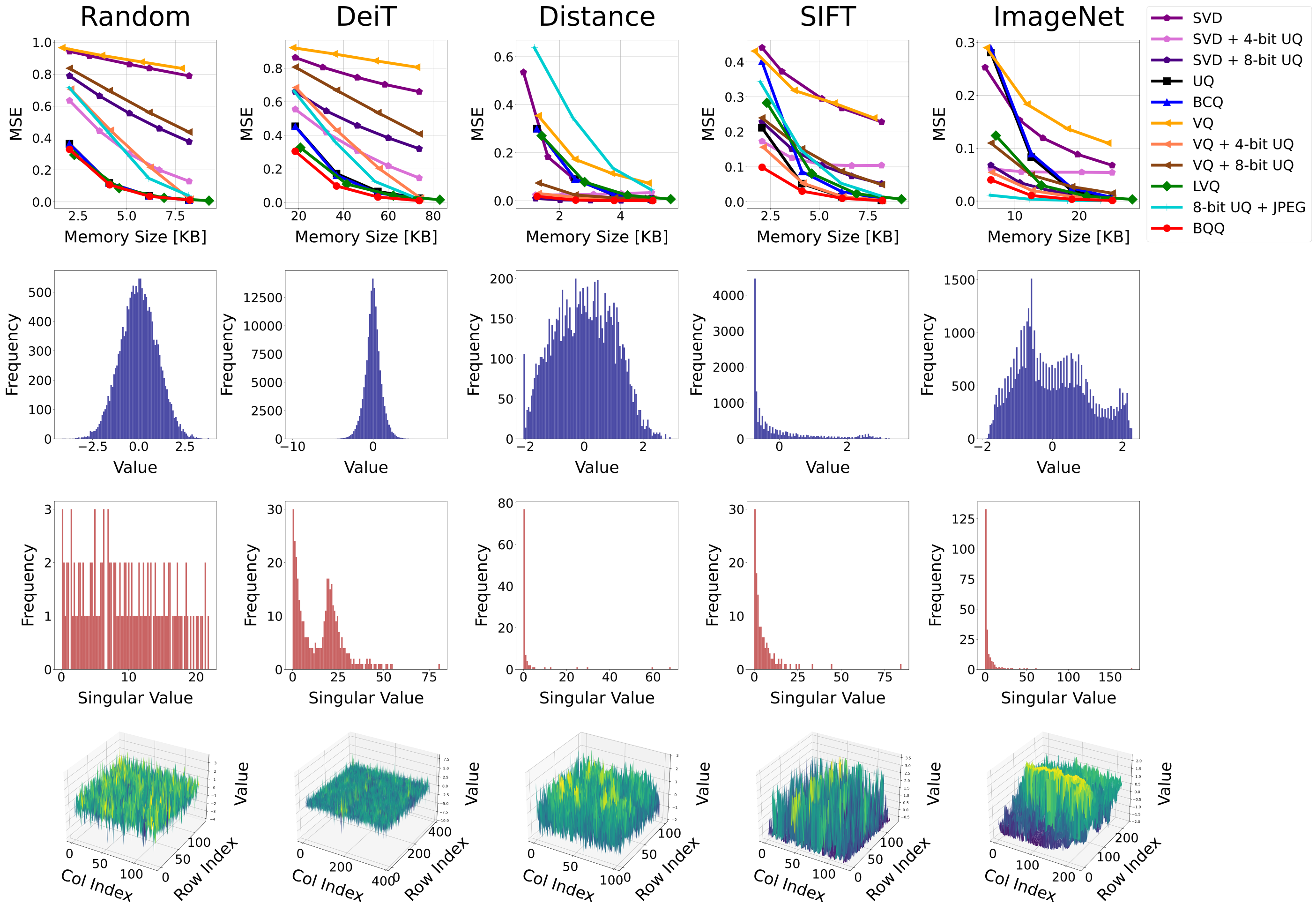}
    \vspace{-0.5em} 
    \caption{\raggedright
    Comparison of the trade-off between reconstruction error (MSE) and memory size for five matrix datasets.
Rows 1–4 show the trade-off curves, value distributions, singular value distributions, and 3D plots of the matrices, respectively.
}
    \label{fig:mse_matrix}
    \vspace{-1.5em} 
\end{figure}

We evaluate the trade-off between approximation error and memory size across five types of real-valued matrices:
(i) a random matrix sampled from a Gaussian distribution,
(ii) a weight matrix from the DeiT-S model~\cite{pmlr-DeiT},
(iii) an inter-city distance matrix from the TSPLIB dataset~\cite{TSPlib},
(iv) a matrix composed of multiple 128-dimensional feature vectors extracted from the SIFT dataset~\cite{jegou_sift}, commonly used in ANN search, 
(v) a red channel matrix of an image from the ImageNet dataset~\cite{ImageNet}.
Each matrix is standardized to have zero mean and a variance of one prior to quantization.
We compare nine methods:
(1) SVD, a low-rank approximation using SVD;
(2) SVD + $p$-bit UQ, SVD low-rank approximation followed by $p$-bit UQ of the factorized matrices;
(3) UQ;
(4) BCQ;
(5)VQ, vector quantization where groups of values are clustered using k-means and encoded as indices of a codebook;
(6) VQ + $p$-bit UQ, VQ followed by $p$-bit UQ of centroids;
(7)$E_8$ LVQ, lattice vector quantization using the $E_8$ lattice with 240 centroids of norm $\sqrt{2}$, representing each 8-dimensional input as a linear combination of centroids stored in 8 bits;
(8) 8-bit UQ + JPEG; a combination of 8-bit UQ and JPEG-style compression~\cite{wallace1991_jpeg}, where discrete cosine transformation is applied before quantization to exploit spatial redundancy in images;
(9) \abbrevmethod.
 The performance is measured in terms of MSE and the memory size of the quantized matrices. 

Fig.~\ref{fig:mse_matrix} presents trade-off curves along with visualizations of value distributions, singular value distributions, and 3D surface plots derived from the original matrices.
Across all datasets, \abbrevmethod consistently achieves a superior trade-off between compression rate and reconstruction accuracy, demonstrating its general effectiveness for matrix data approximation.
Notably, the advantage of \abbrevmethod over UQ, BCQ, and LVQ becomes especially pronounced for matrices whose singular value spectrum is dominated by a few large components. In contrast, for matrices with relatively flat singular value distributions—i.e., those lacking dominant components—the gain over them is smaller. This suggests that \abbrevmethod particularly benefits from matrices with concentrated spectral energy.
On the other hand, when compared to SVD, SVD + UQ, VQ, and VQ + UQ, the opposite trend is observed: \abbrevmethod exhibits greater advantage for matrices with more uniform singular value spectra.
Admittedly, on the ImageNet dataset, \abbrevmethod yields a less favorable trade-off between memory size and reconstruction error compared to JPEG. However, since JPEG combines discrete cosine transform with quantization, integrating \abbrevmethod with transform-based approaches could potentially lead to further improvements in compression efficiency.

\vspace{-1em}
\paragraph{Post-Training Quantization for ViTs}\label{subsec:post_training_quantization}
Following the methodology outlined in Sec.~\ref{subsec:ptq_method}, we evaluate the performance of \abbrevmethod on pretrained ViTs. Specifically, we compare \abbrevmethod with several leading PTQ methods, including COMQ~\cite{zhang2024_COMQ}, FQ-ViT~\cite{ijcai2022p164_FQ-ViT}, PTQ4ViT~\cite{PTQ4ViT}, RepQ-ViT~\cite{Li_2023_ICCV_RepQ-ViT}, ERQ~\cite{ICML_ERQ}, and PSAQ-ViT~\cite{PSAQ-ViT1}, using two representative ViT architectures: DeiT~\cite{pmlr-DeiT} and Swin Transformer~\cite{Liu2021_Swin}. All methods are tested under the 32-bit activation setting to ensure a fair comparison.
Additionally, to appropriately assess the effectiveness of \abbrevmethod, we also compare it with variants that use the exact same process but replace the quantization method with UQ or BCQ, as baselines.
In our quantization framework, we apply group-wise quantization with submatrices of size $384 \times 384$ for DeiTs and $96 \times 96$ for Swins based on the first block's channel number. If a weight matrix matches the group size, it is not subdivided further (i.e., layer-wise quantization). As a special case, the final classification layer uses a group size of $100 \times 96$. Also, the linear patch embedding layer in DeiT is grouped based on its embedding dimension ($384 \times 384$ for DeiT-S and $784 \times 784$ for DeiT-B), while Swin's patch embedding layer, implemented as a convolutional layer, uses channel-wise UQ instead of \abbrevmethod.
The same group sizes are applied in the UQ and BCQ baselines for fair comparison.
In the case of bias and normalization parameter correction (denoted as c-UQ, c-BCQ, and c-BQQ for each quantization method), we optimize them using the Adam optimizer~\cite{2015-kingma_ADAM} with a learning rate of 0.001 for 15 epochs via a minibatch size of 16, and calibration data are randomly selected from the ImageNet~\cite{ImageNet} training dataset, with 2048 samples for DeiTs and 1024 samples for Swins, in accordance with~\cite{zhang2024_COMQ} setting.

\begin{table}[t]
\vspace{-2.5em} 
\caption{Comparison of ImageNet top-1 accuracy across various quantization methods on ViTs.}\label{tab:accuracy}
\vspace{0.5em} 
\renewcommand{\arraystretch}{1.15} 
\centering
\fontsize{8pt}{8pt}\selectfont
\begin{tabular}{ccccrrrr}
\hline
\multirow{2}{*}{\textbf{Method}} & \multirow{2}{*}{\textbf{W bit}} & \multirow{2}{*}{\textbf{Data Free}} & \multirow{2}{*}{\textbf{W scale}} & \multicolumn{4}{c}{\textbf{Top-1 Accuracy {[}\%{]}}}                                                                                                  \\
                                 &                                 &                                     &                                     & \multicolumn{1}{c}{\textbf{DeiT-S}} & \multicolumn{1}{c}{\textbf{DeiT-B}} & \multicolumn{1}{c}{\textbf{Swin-T}} & \multicolumn{1}{c}{\textbf{Swin-S}} \\ \hline
\textbf{COMQ}                    & 2                               & $\times$                                   & column-wise                         & 67.19                               & 77.14                               & \textbf{74.05}                      & 78.02                               \\
\textbf{ERQ}                     & 2                               & $\times$                                   & column-wise                         & 31.95                               & 63.67                               & 45.97                               & 35.44                               \\
\textbf{RepQ-ViT}                & 2                               & $\times$                                   & column-wise                         & 0.31                                & 0.42                                & 0.12                                & 0.12                                \\
\textbf{c-UQ}                    & 2                               & $\times$                                   & group-wise                          & 52.21                               & 60.57                               & 67.49                               & 74.16                               \\
\textbf{c-BCQ}                   & 2                               & $\times$                                   & group-wise                          & 60.13                               & 73.37                               & 68.09                               & 73.97                               \\
\textbf{c-BQQ}                   & 2*                              & $\times$                                   & group-wise                          & \textbf{69.41}                      & \textbf{77.94}                      & 74.03                               & \textbf{78.47}                      \\ \hline
\textbf{PSAQ-ViT}                & 2                               & $\checkmark$                                   & column-wise                                   & 0.27                                & 0.19                                & 0.15                                & 0.14                                \\
\textbf{UQ}                      & 2                               & $\checkmark$                                   & group-wise                          & 3.23                                & 2.45                                & 14.69                               & 30.69                               \\
\textbf{BCQ}                     & 2                               & $\checkmark$                                   & group-wise                          & 10.83                               & 12.99                               & 18.62                               & 34.84                               \\
\textbf{BQQ}                     & 2*                              & $\checkmark$                                   & group-wise                          & \textbf{58.25}                      & \textbf{72.09}                      & \textbf{57.37}                      & \textbf{68.17}                      \\ \hline
\textbf{COMQ}                    & 3                               & $\times$                                   & column-wise                         & \textbf{77.47}                      & 80.47                               & 79.31                               & \textbf{81.95}                      \\
\textbf{ERQ}                     & 3                               & $\times$                                   & column-wise                         & 75.56                               & 79.73                               & 77.99                               & 80.87                               \\
\textbf{RepQ-ViT}                & 3                               & $\times$                                   & column-wise                         & 58.26                               & 68.80                               & 21.41                               & 69.57                               \\
\textbf{FQ-ViT}                  & 3                               & $\times$                                   & column-wise                         & 51.06                               & 65.64                               & 65.38                               & 71.88                               \\
\textbf{PTQ4ViT}                 & 3                               & $\times$                                   & layer-wise                          & 70.22                               & 75.42                               & 70.74                               & 73.46                               \\
\textbf{c-UQ}                    & 3                               & $\times$                                   & group-wise                          & 72.08                               & 78.85                               & 78.11                               & 80.91                               \\
\textbf{c-BCQ}                   & 3                               & $\times$                                   & group-wise                          & 75.53                               & 79.78                               & 78.60                               & 81.19                               \\
\textbf{c-BQQ}                   & 3*                              & $\times$                                   & group-wise                          & 77.33                               & \textbf{80.81}                      & \textbf{79.34}                      & 81.86                               \\ \hline
\textbf{PSAQ-ViT}                & 3                               & $\checkmark$                                   & column-wise                                   & 52.76                               & 66.40                               & 65.87                               & 72.53                               \\
\textbf{UQ}                      & 3                               & $\checkmark$                                   & group-wise                          & 42.28                               & 58.56                               & 70.90                               & 75.72                               \\
\textbf{BCQ}                     & 3                               & $\checkmark$                                   & group-wise                          & 63.46                               & 69.09                               & 72.99                               & 76.18                               \\
\textbf{BQQ}                     & 3*                              & $\checkmark$                                   & group-wise                          & \textbf{75.61}                      & \textbf{79.90}                      & \textbf{77.33}                      & \textbf{80.36}                      \\ \hline
\textbf{COMQ}                    & 4                               & $\times$                                   & column-wise                         & 78.98                               & 81.40                               & \textbf{80.89}                      & 82.85                               \\
\textbf{ERQ}                     & 4                               & $\times$                                   & column-wise                         & 78.95                               & 81.46                               & 80.85                               & \textbf{82.99}                      \\
\textbf{RepQ-ViT}                & 4                               & $\times$                                   & column-wise                         & 75.39                               & 78.77                               & 75.08                               & 81.53                               \\
\textbf{FQ-ViT}                  & 4                               & $\times$                                   & column-wise                         & 76.23                               & 79.92                               & 78.81                               & 81.89                               \\
\textbf{PTQ4ViT}                 & 4                               & $\times$                                   & layer-wise                          & 77.50                               & 80.07                               & 78.46                               & 80.24                               \\
\textbf{c-UQ}                    & 4                               & $\times$                                   & group-wise                          & 78.15                               & 81.01                               & 80.42                               & 82.40                               \\
\textbf{c-BCQ}                   & 4                               & $\times$                                   & group-wise                          & 78.67                               & 81.22                               & 80.46                               & 82.47                               \\
\textbf{c-BQQ}                   & 4*                              & $\times$                                   & group-wise                          & \textbf{79.12}                      & \textbf{81.47}                      & 80.57                               & 82.72                               \\ \hline
\textbf{PSAQ-ViT}                & 4                               & $\checkmark$                                   & column-wise                                   & 76.59                               & 80.23                               & 79.15                               & 81.94                               \\
\textbf{UQ}                      & 4                               & $\checkmark$                                   & group-wise                          & 73.53                               & 77.49                               & 79.17                               & 81.47                               \\
\textbf{BCQ}                     & 4                               & $\checkmark$                                   & group-wise                          & 75.82                               & 78.58                               & 79.63                               & 81.72                               \\
\textbf{BQQ}                     & 4*                              & $\checkmark$                                   & group-wise                          & \textbf{78.76}                      & \textbf{81.20}                      & \textbf{80.21}                      & \textbf{82.21}                      \\ \hline
\textbf{Full Precision}          & 32                              & -                                   & -                                   & 79.83                               & 81.80                               & 81.37                               & 83.21                               \\ \hline
\end{tabular}
\begin{center}
    \footnotesize
$\ast$: Pseudo $p^*$-bit \abbrevmethod has a model size matching that of a $p$-bit quantized model, despite each matrix being 1 bit.

\end{center}
\vspace{-2.5em} 
\end{table}
\normalsize

Tab.~\ref{tab:accuracy} summarizes the experimental results. 
Here, \textbf{W bit} denotes the bit width for weight quantization, while \textbf{W scale} indicates the granularity of scaling factors (e.g., per-layer, per-group, or per-column).
 Note that although each weight matrix in \abbrevmethod is binary, its configuration is designed to match the information content of a first-order $p$-bit quantized model, which we refer to as pseudo $p^*$-bit.
The results for COMQ~\cite{zhang2024_COMQ}, FQ-ViT~\cite{ijcai2022p164_FQ-ViT}, and PTQ4ViT~\cite{PTQ4ViT} are cited from~\cite{zhang2024_COMQ}, while those for RepQ-ViT~\cite{Li_2023_ICCV_RepQ-ViT}, ERQ~\cite{ICML_ERQ}, and PSAQ-ViT~\cite{PSAQ-ViT1} were obtained using publicly available implementations.
As shown in the experimental results, \abbrevmethod consistently achieves SOTA performance regardless of whether calibration data is used, demonstrating a compelling trade-off between accuracy and model size. In particular, it shows notable improvements both in data-free settings and in configurations with model size equivalent to 2-bit quantization.  
To the best of our knowledge, this is the first study to achieve practically usable accuracy with a model size equivalent to 2-bit quantization in the absence of any data.  
In addition, while most existing methods preserve accuracy by using column-wise scaling factors--resulting in larger model size--\abbrevmethod adopts group-wise scaling, which can reduce parameter overhead. Despite using a more compact scaling strategy, it still achieves competitive accuracy.

\vspace{-0.7ex}
\section{Discussion}\label{sec:discussion}
\vspace{-0.5ex}
\paragraph{\abbrevmethod Effectiveness and Characteristics}
As shown in the matrix compression experiments, the advantage of \abbrevmethod over UQ, BCQ, and LVQ becomes more pronounced for matrices with skewed singular value distributions, where a few dominant singular values capture most of the spectral energy. Conversely, when the singular values are more uniformly distributed, the performance gap between these methods and \abbrevmethod becomes smaller.
On the other hand, \abbrevmethod shows a clear advantage over SVD- and VQ-based methods when the singular value spectrum is relatively flat. This is likely because SVD and VQ are designed to capture and compress redundant patterns in the matrix, which works well for low-rank or structured data. When such redundancy is absent—as in random-like matrices with weak spectral bias—their performance tends to degrade.
Unlike SVD and VQ, UQ, BCQ, and LVQ quantize each element or vector independently to its nearest representative value. This makes it difficult to exploit pattern redundancy, but it also allows these methods to remain relatively stable across different spectral shapes. In fact, when the data lacks significant structure, such independent quantization can lead to more efficient compression than pattern-based approaches.
Overall, \abbrevmethod integrates the strengths of both pattern-oriented and element-wise quantization strategies. It leverages the ability to capture structural redundancy—similar to SVD and VQ—while also benefiting from the stability and granularity of scalar quantization methods like UQ, BCQ, and LVQ. As a result, it achieves robust compression performance across a wide range of singular value distributions.

In addition, PTQ experiments on ViTs demonstrate that \abbrevmethod\ achieves SOTA performance in both data-free and calibration-based settings. While COMQ or ERQ slightly outperforms it in some cases, they adopt channel-wise quantization with more scaling parameters, whereas our group-wise approach yields a more compact model. Moreover, in contrast to most PTQ methods that optimize discrete parameters by minimizing output error, our approach optimizes binary parameters by simply minimizing the reconstruction error from the original weight matrix (i.e, no PTQ-specific binary variable optimization is performed). 
Despite this, \abbrevmethod\ matches or even surpasses PTQ-specialized methods, which is a noteworthy outcome.
These results are likely due to \abbrevmethod's ability to capture structural redundancy often overlooked by first-order methods in overparameterized layers. 
It is also noteworthy that the matrix multiplication between weights and inputs can be performed using only addition operations, resulting in minimal computational overhead for inference (see App.\ref{subsec:appendix_inference_cost}).
This suggests that significant acceleration could be achieved with specialized hardware.

\paragraph{Further Potential and Limitations}
Despite the demonstrated effectiveness of \abbrevmethod, there remains significant room for further improvement. In the current implementation, we adopt a greedy optimization strategy as described in Alg.~\ref{alg:greedy_qq}, which is suboptimal from a global perspective. Jointly optimizing all binary matrices and scaling factors could potentially lead to further reductions in quantization error.
In addition, although our PTQ framework with \abbrevmethod is based on minimizing weight approximation error--except for the correction of bias and normalization parameters--it is generally more effective to minimize output quantization error, as demonstrated in many previous studies. Adapting \abbrevmethod to optimize binary matrices with respect to output error could therefore lead to even greater PTQ accuracy. Nevertheless, it is noteworthy that \abbrevmethod already achieves SOTA performance.
Moreover, in our experiments, the intermediate dimension is fixed, as described in Sec.~\ref{subsec:implemantation_detail}. However, this configuration may not be optimal under a fixed binary parameter budget. Exploring the optimal ratio between the intermediate dimension and the number of binary matrix stacks ($p$ in Eq.~\eqref{eq:qq_approx}) could further improve approximation error (see App.~\ref{app:error-stack-width}).
While \abbrevmethod\ holds such potential, this work has certain limitations. 
Specifically, while an upper bound on the approximation error is provided (see App.~\ref{app:approx_bound}), it does not yet establish a theoretical guarantee that our method outperforms first-order quantization under specific conditions.
Additionally, the quantization process still incurs a non-negligible computational cost (see App.~\ref{app:bqq_execution_time_for_ptq}).
Nonetheless, we believe that \abbrevmethod\ has the capacity to contribute to a wide range of applications beyond the experiments presented in this study, and could have a significant impact.
\vspace{-0.25em}
\vspace{-0.5ex}
\section{Conclusion}
\vspace{-0.7ex}
We introduced \method (\abbrevmethod), a novel quantization framework that approximates real-valued matrices as linear combinations of binary matrix products. 
Across both matrix compression and ViT-based PTQ tasks, \abbrevmethod\ consistently outperforms existing methods in terms of accuracy and compression ratio.
These findings highlight the remarkable capability of second-order binary representations in capturing complex structures beyond the reach of first-order schemes, while maintaining an extremely compact data format. 
By providing an expressive and versatile framework for compressing real-valued matrices using binary bases, \abbrevmethod opens new possibilities for building efficient, scalable systems across a wide range of machine learning and information processing applications. We believe this work lays the groundwork for future research into quadratic binary representations and their role in high-performance model compression, retrieval systems, and large-scale learning on massive training data.

\section*{Acknowledgements}
This work was supported by JST-ALCA-Next (Grant JPMJAN24F3),
by JST PRESTO (Grant JPMJPR23P1),  
and by JST SPRING (Grant JPMJSP2180).

\bibliography{bibsample}  
\bibliographystyle{plainnat}
\clearpage

\appendix
\section*{Appendix}
\renewcommand{\thefigure}{S.\arabic{figure}}
\renewcommand{\thetable}{S.\arabic{table}}
\renewcommand{\theequation}{S.\arabic{equation}}
\renewcommand{\thealgorithm}{S.\arabic{algorithm}}  

\setcounter{figure}{0}
\setcounter{table}{0}
\setcounter{equation}{0}
\setcounter{algorithm}{0}

\renewcommand{\thesection}{A}
\renewcommand{\thesubsection}{A.\arabic{subsection}}
\setcounter{subsection}{0}  

\subsection{Derivation of Equation~\eqref{eq:KL_pubo}}\label{app:appendix_kl_pubo_proof}

In order to extend the QUBO formulation, which is the domain of application for AMFD, to a general PUBO formulation, we prove that the KL divergence between the mean-field approximate distribution and the canonical distribution for the PUBO formulation is given by Eq.~\eqref{eq:KL_pubo}.

\begin{proof}
\label{pr:KL_mean_field}
$\quad\\$
\vspace{2em}
From the definition of KL divergence:
\small
\begin{align}
    D_\text{KL}\left(P_\text{MF}\left(\boldsymbol{s}\right) \mid\mid P_\text{C}\left(\boldsymbol{s}\right)\right)
    &= \sum_{\boldsymbol{s}} P_\text{MF}\left(\boldsymbol{s}\right) 
    \ln\left(\frac{P_\text{MF}\left(\boldsymbol{s}\right)}{P_\text{C}\left(\boldsymbol{s}\right)}\right) \notag\\
    &= \sum_{\boldsymbol{s}} \left[ 
        \left(\prod_{i=1}^{N} p_i\left(s_i\right)\right) 
        \left(
        \frac{1}{T} L\left(\boldsymbol{s}\right) 
            + \ln\left(Z\right)
            + \sum_{i=1}^{N} \ln\left(p_i\left(s_i\right)\right) 
        \right)
    \right].
    \label{eq:KL_definition_PUBO}
\end{align}
\normalsize

For the first term in Eq.~\eqref{eq:KL_definition_PUBO}:
\small
        \begin{align}
            &\sum_{\boldsymbol{s}} \left[\left(\prod_{j=1}^{N} p_j\left(s_j\right)\right)L(\boldsymbol{s})\right]\notag\\
            &=\sum_{\boldsymbol{s}} \left[\left(\prod_{j=1}^{N} p_j\left(s_j\right)\right)\left(\sum_{i_1}J_{i_1}^{(1)}s_i +\sum_{i_1<i_2}J_{i_1 i_2}^{(2)}s_{i_1}s_{i_2} + \sum_{i_1<i_2<i_3} J_{i_1 i_2 i_3}^{(3)}s_{i_1}s_{i_2}s_{i_3} + \cdot\cdot\cdot\right)\right]\notag\\
            &= \sum_{\boldsymbol{s}} \left(\prod_{j\ne i_1} p_j\left(s_j\right)\sum_{i_1}J_{i_1}^{(1)}s_{i_1} p_{i_1}\left(s_{i_1}\right)\right)
            +\sum_{\boldsymbol{s}} \left(\prod_{j\ne i_1,i_2} p_j\left(s_j\right)\sum_{i_1<i_2}J_{i_1 i_2}^{(2)}s_{i_1}p_{i_1}\left(s_{i_1}\right)s_{i_2}p_{i_2}\left(s_{i_2}\right)\right)\notag\\ 
            &+\sum_{\boldsymbol{s}}\left(\prod_{j\ne i_1,i_2,i_3} p_j\left(s_j\right)\sum_{i_1<i_2<i_3}J_{i_1 i_2 i_3}^{(3)}s_{i_1}p_{i_1}\left(s_{i_1}\right)s_{i_2}p_{i_2}\left(s_{i_2}\right)s_{i_3}p_{i_3}\left(s_{i_3}\right)\right) + \cdot\cdot\cdot\notag\\
            &=\left(\prod_{j\ne i_1}\sum_{s_j=0}^{1}p_j\left(s_j\right)\right)
            \sum_{i_1}J_{i_1}^{(1)}\sum_{s_{i_1}=0}^{1}s_{i_1}p_{i_1}\left(s_{i_1}\right) \notag\\        
            &+\left(\prod_{j\ne i_1,i_2}\sum_{s_j=0}^{1}p_j\left(s_j\right)\right)\left(\sum_{i_1<i_2}J_{i_1 i_2}^{(2)}\sum_{s_{i_1}=0}^{1}s_{i_1}p_{i_1}\left(s_{i_1}\right)\sum_{s_{i_2}=0}^{1}s_{i_2}p_{i_2}\left(s_{i_2}\right)\right)\notag\\
            &+
            \left(\prod_{j\ne i_1,i_2,i_3}\sum_{s_j=0}^{1}p_j\left(s_j\right)\right)\left(\sum_{i_1<i_2<i_3}J_{i_1 i_2 i_3}^{(3)}\sum_{s_{i_1}=0}^{1}s_{i_1}p_{i_1}\left(s_{i_1}\right)\sum_{s_{i_2}=0}^{1}s_{i_2}p_{i_2}\left(s_{i_2}\right) \sum_{s_{i_3}=0}^{1}s_{i_3}p_{i_3}\left(s_{i_3}\right)\right)\cdot\cdot\cdot\notag\\
            &=\sum_{i=1}^{N}J_{i_1}^{(1)}x_i+\sum_{i=1}^{N}\sum_{j<i}J_{i_1 i_2}^{(2)}x_ix_j + \sum_{i=1}^{N}\sum_{j<i}\sum_{k<j}J_{i_1 i_2 i_3}^{(3)}x_ix_jx_k \cdot\cdot\cdot
            + \sum_{i_1<i_2<\cdots<i_k} J_{i_1i_2\cdots i_k}^{(k)} \prod_{j=1}^{k} x_{i_j}\notag\\
            \label{eq:mean_field_energy}
            & = L(\boldsymbol{x}).
        \end{align}
\normalsize

For the second term in Eq.~\eqref{eq:KL_definition_PUBO}:
\small
\begin{align}
            &\sum_{\boldsymbol{s}} \left(\prod_{i=1}^{N} p_i\left(s_i\right)\right)\ln(Z)\notag\\
            &=\left(\prod_{i=1}^{N}\sum_{s_i=0}^{1}p_i\left(s_i\right)\right)\ln(Z)\notag\\
            \label{eq:free_energy}
            &= \ln(Z) \quad  \left( \because \sum_{s_j=0}^{1}p_j\left(s_j\right)=1\right).
        \end{align}
\normalsize

For the third term in Eq.~\eqref{eq:KL_definition_PUBO}:
\small
             \begin{align}
            &\sum_{\boldsymbol{s}} \left[\left(\prod_{i=1}^{N} p_i\left(s_i\right)\right)\sum_{i=1}^{N} \ln\left(p_i\left(s_i\right)\right)
            \right]\notag\\
            &= \sum_{\boldsymbol{s}}            
                \left[
                    \sum_{i=1}^{N} \left(\prod_{j\ne i}p_j\left(s_j\right)\right)p_i\left(s_i\right)\ln\left(p_i\left(s_i\right)\right) 
                \right]\notag\\
            &=\sum_{i=1}^{N} \left(\prod_{j\ne i}\sum_{s_j=0}^{1}p_j\left(s_j\right)\right)\sum_{s_i=0}^{1}p_i\left(s_i\right)\ln\left(p_i\left(s_i\right)\right) \notag\\ 
            &=\sum_{i=1}^{N}\sum_{s_i=0}^{1}p_i\left(s_i\right)\ln\left(p_i\left(s_i\right)\right) \quad \left(\because \sum_{s_j=0}^{1}p_j\left(s_j\right)=1\right)\notag \\
            \label{eq:mean_field_entropy}
            &=\sum_{i=1}^{N}\left[x_i\ln x_i+(1-x_i)\ln (1-x_i)\right]
            \quad \left(\because x_i = \sum_{s_i=0}^{1}s_ip_i(s_i)=p_i(1)\right).
        \end{align}
\normalsize

From Eq.~\eqref{eq:KL_definition_PUBO}--\eqref{eq:mean_field_entropy}, we obtain the KL divergence expression in Eq.~\eqref{eq:KL_pubo}. Therefore, the claim is proven.
\end{proof}

\subsection{Uniform Quantization with Grid Search}\label{subsec:appendix_uq_alg}

Here, we provide a detailed explanation of UQ algorithm introduced in Sec.~\ref{sec:evaluation}. As shown in Alg.~\ref{alg:UQ_grid_search}, the scaling factor is determined via grid search so as to minimize the MSE between the original matrix and the quantized matrix. 
We optimized with $N_\text{split}=100$ for all experiments.

\begin{algorithm}[h]
\caption{MSE-Aware Uniform Quantization with Grid Search}
\label{alg:UQ_grid_search}
\begin{algorithmic}[1]
\Require Input matrix $\boldsymbol{W}$, bit width $N_{\text{bit}}$, the number of grid divisions $N_\text{split}$
\Ensure Quantized matrix $\boldsymbol{Q}$, scaling factor $a$, bias $b$, dequantized matrix $\boldsymbol{W}_q$
\State $L \gets 2^{N_{\text{bit}}}$ \Comment{The number of quantization levels}
\State $\mu \gets \text{mean}(\boldsymbol{W})$ \Comment{Search range setting}
\State $w_{\min} \gets \text{min}(\boldsymbol{W})$
\State $w_{\max} \gets \text{max}(\boldsymbol{W})$
\State Initialize $\varepsilon \gets \infty$, $\boldsymbol{Q} \gets \text{None}$, $(\alpha, \beta) \gets \text{None}$
\For{each $r_{\max} \in \text{linspace}(\mu, w_{\max}, N_\text{split})$}
    \For{each $r_{\min} \in \text{linspace}(w_{\min}, \mu, N_\text{split})$}
        \State $\boldsymbol{W}_c \gets \text{clip}(\boldsymbol{W}, r_{\min}, r_{\max})$
        \Comment{Clip $\boldsymbol{W}$ to $[r_{\min}, r_{\max}]$}

        \State 
        $
        \boldsymbol{Q}_c \gets \left\lfloor \frac{\boldsymbol{W}_c - r_{\min}}{r_{\max} - r_{\min}} \cdot (L - 1)  \right\rceil
        $
        \Comment{Quantize to $[0, L-1]$}
        \State 
        $
        \boldsymbol{W}_c \gets \frac{\boldsymbol{Q}_c}{L - 1} \cdot (r_{\max} - r_{\min}) + r_{\min}
        $
        \Comment{Dequantize}
        \State 
        $
        e \gets \frac{1}{|\boldsymbol{W}|} \sum (\boldsymbol{W} - \boldsymbol{W}_{c})^2
        $
        \Comment{Compute quantization error}
        \If{$e < \varepsilon$}
            \State $\varepsilon \gets e$ \Comment{Save best quantization setting}
            \State $\boldsymbol{Q} \gets \boldsymbol{Q}_c$
            \State $a \gets \frac{r_{\max} - r_{\min}}{L - 1}$, $b \gets r_{\min}$
            \State $\boldsymbol{W}_q \gets \boldsymbol{W}_c$
        \EndIf
    \EndFor
\EndFor
\Return $\boldsymbol{Q}$, $a$, $b$,  $\boldsymbol{W}_q$
\end{algorithmic}
\end{algorithm}

\subsection{$E_8$ Lattice Vector Quantization using 240 centroids of norm $\sqrt{2}$}\label{subsec:appendix_lvq_alg}

Here, we describe the $E_8$ LVQ algorithm introduced in Sec.~\ref{sec:evaluation}. 
In this work, we construct a codebook consisting of all $240$ $E_8$ lattice vectors with Euclidean norm $\sqrt{2}$ as centroids. 
For a given 8-dimensional vector, the nearest centroid is selected to approximate it. 
To minimize the MSE, after selecting the centroid based on cosine similarity, the optimal scalar scaling factor is computed. 
Each 8-dimensional vector is thus associated with a single 8-bit index, which is equivalent to approximately one bit of scalar quantization. 
By iteratively applying this procedure to the residual errors, the quantization can be extended to multiple bits. 
However, maintaining a separate scaling factor for each 8-dimensional vector would make the memory overhead non-negligible. Therefore, in this work, the scaling factors are quantized using 2-bit uniform quantization, which was empirically found to achieve the best trade-off between memory efficiency and quantization error.
The resulting residual-based multi-bit LVQ algorithm is summarized in Alg.~\ref{alg:e8-lvq-matrix}.

\begin{algorithm}[H]
\caption{$E_8$ Lattice Vector Quantization with Residual Greedy Search}
\label{alg:e8-lvq-matrix}
\begin{algorithmic}[1]
\Require Input matrix $\mathbf{W} \in \mathbb{R}^{M \times N}$, number of quantization bits $N_{\text{bits}}$, scale bits $s_{\text{bits}}$
\Ensure Quantized and reconstructed matrix ${\mathbf{W_q}}$, code vector indices $\mathbf{k^*}$, scaling factors $\boldsymbol{\alpha}$

\State Flatten $\mathbf{W}$ to $\mathbf{W}_{\text{flat}}$
\State Pad $\mathbf{W}_{\text{flat}}$ with zeros so that its length is a multiple of $8$
\State Reshape $\mathbf{W}_{\text{flat}}$ to $\mathbf{D} \in \mathbb{R}^{n \times 8}$

\State Construct the $E_8$ codebook $\mathbf{C} \in \mathbb{R}^{240 \times 8}$ consisting of all lattice vectors with norm $\sqrt{2}$

\State Initialize reconstruction $\mathbf{D}_{\text{total}} \gets 0$

\For{$\text{bit} = 1$ to $N_{\text{bits}}$}
    \State Normalize codebook and data: $\mathbf{C}_{\text{norm}} = \mathbf{C}/\|\mathbf{C}\|_2$, $\mathbf{D}_{\text{norm}} = \mathbf{D}/\|\mathbf{D}\|_2$
    \State Compute cosine similarity matrix: 
    \[
        \mathbf{S} = \mathbf{D}_{\text{norm}} \mathbf{C}_{\text{norm}}^\top
    \]
    \State Select closest code vector index for each row: 
    \[
        k^*_r = \arg\max_{k=1,\dots,240} S_{r,k}, \quad r=1,\dots,n
    \]
    \State Select corresponding codes: $\mathbf{C}_{\text{selected}} \gets \mathbf{C}[k^*]$
    
    \State Compute scalar coefficients $\boldsymbol{\alpha} \in \mathbb{R}^n$ for each row of $\mathbf{D}$:
    $\displaystyle \alpha_r = \frac{\sum_{j=1}^{8} D_{r,j} \, C_{\text{selected}, r, j}}{\sum_{j=1}^{8} C_{\text{selected}, r, j}^2}$
    \State Quantize $\boldsymbol{\alpha}$ using uniform quantization with $s_{\text{bits}}$
    \State Reconstruct partial matrix: $\mathbf{D}_{\text{hat}} \gets \boldsymbol{\alpha} \cdot \mathbf{C}_{\text{selected}}$
    \State Update residual: $\mathbf{D} \gets \mathbf{D} - \mathbf{D}_{\text{hat}}$
    \State Accumulate reconstruction: $\mathbf{D}_{\text{total}} \gets \mathbf{D}_{\text{total}} + \mathbf{D}_{\text{hat}}$
\EndFor

\State Reshape $\mathbf{D}_{\text{total}}$ to original shape of $\mathbf{W}$: ${\mathbf{W_q}} \gets \mathbf{D}_{\text{total}}$
\State \Return ${\mathbf{W_q}}, \mathbf{k^*}, \boldsymbol{\alpha}$
\end{algorithmic}
\end{algorithm}

\subsection{Inference Computational Cost Analysis of the BQQ Layer in DNN}\label{subsec:appendix_inference_cost}

Since $p$-bit quantization typically incurs approximately $p$ times the computational cost of 1-bit quantization, we analyze the computational cost based on the 1-bit case for both the conventional first-order quantization and the proposed BQQ layer.
We focus on a single linear layer, which is where weights in ViTs are concentrated.
 Let the input be a real-valued matrix \( \mathbf{X} \in \mathbb{R}^{n \times d} \), and the 1-bit quantized weight for the first-order baseline be \( \mathbf{W}_q \in \{0, 1\}^{m \times n} \). For the BQQ method, the weights are represented as \( \mathbf{Y} \in \{0, 1\}^{m \times l} \) and \( \mathbf{Z} \in \{0, 1\}^{l \times n} \), where $l$ is the intermediate rank.

Here, \texttt{AND} refers to bitwise operations between binary weights and real-valued inputs (not binary-binary operations), \texttt{ADD} denotes real-valued addition, and \texttt{MUL} represents real-valued multiplication.
 As described in the experimental setting in Sec.~\ref{subsec:implemantation_detail}, we set $l = \text{round}\left(\frac{mn}{m + n}\right)$ for the BQQ layer. Note that the cost of computing the zero-point bias is omitted because it is a common term and does not change as the bit width increases.

\paragraph{First-Order 1-Bit Quantization}
\[
\text{OUT}_\text{FOQ} = a \mathbf{W}_q \mathbf{X}
\]
The matrix multiplication involves the following computational cost:
\[
\text{COST}_\text{FOQ} = mnd\, \texttt{AND} + md(n-1)\, \texttt{ADD} + md\, \texttt{MUL}
\]

\paragraph{BQQ 1*-Bit Quantization}
The output is computed as:
\[
\text{OUT}_\text{BQQ} = \left( r \mathbf{Y} \mathbf{Z} + s \mathbf{Y} \mathbf{1}_Z + t \mathbf{1}_Y \mathbf{Z} \right) \mathbf{X}
\]
The computational steps are broken down as follows:
\begin{align*}
\mathbf{A} &\leftarrow \mathbf{Z} \mathbf{X} : lnd\, \texttt{AND} + (n-1)ld\, \texttt{ADD} \\
\mathbf{B} &\leftarrow \mathbf{1}_Z \mathbf{X} : (n-1)d\, \texttt{ADD} \\
\mathbf{C} &\leftarrow \mathbf{Y}(r\mathbf{A} + s\mathbf{B}) : 2ld\, \texttt{MUL} + ld\, \texttt{ADD} + mld\, \texttt{AND} + (l-1)md\, \texttt{ADD} \\
\mathbf{D} &\leftarrow \mathbf{1}_Y \cdot t\mathbf{A} : (l-1)d\, \texttt{ADD} + ld\, \texttt{MUL} \\
\text{Out} &\leftarrow \mathbf{C} + \mathbf{D} : md\, \texttt{ADD}
\end{align*}

The total computational cost becomes:
\[
\text{COST}_\text{BQQ} = ld(m + n)\, \texttt{AND} + d[(m+n+1)l + n - m - 2]\, \texttt{ADD} + 3ld\, \texttt{MUL}
\]

\paragraph{Relative Cost Ratio}
We compare the computational cost between the first-order 1-bit quantization and the BQQ method. The relative ratio is given by:
\[
\frac{\text{COST}_\text{BQQ}}{\text{COST}_\text{FOQ}}=\frac{
    ld(m + n)\, \texttt{AND} + d[(m+n+1)l + n - m - 2]\, \texttt{ADD} + 3ld\, \texttt{MUL}
}{
    mnd\, \texttt{AND} + md(n-1)\, \texttt{ADD} + md\, \texttt{MUL}
}
\]
\[
= 1 + \frac{d[(n + l - 2)\, \texttt{ADD} + (3l - m)\, \texttt{MUL}]}{mnd\, \texttt{AND} + md(n - 1)\, \texttt{ADD} + md\, \texttt{MUL}},  \quad \left(\text{where } l = \frac{mn}{m+n}\right)
\]
yielding a computational complexity ratio of $\mathcal{O}(1)$, since the first-order quantization has 
$\mathcal{O}(mnd)$ operations while the BQQ method has $\mathcal{O}((m+n)ld)$ operations, and for 
$l \approx \frac{mn}{m+n}$, their ratio becomes of order one:
\[
\frac{\text{COST}_\text{BQQ}}{\text{COST}_\text{FOQ}}
= \mathcal{O}\!\left(\frac{(m+n)ld}{mnd}\right)
= \mathcal{O}(1), \quad \text{when } l \approx \frac{mn}{m+n}.
\]

\paragraph{Practical Examples}
For the DeiT model in Sec.~\ref{subsec:post_training_quantization} with \( m = n = 384 \), \( l = 192 \):
\[
1 + \frac{574\, \texttt{ADD} + (3 \cdot 192 - 384)\, \texttt{MUL}}{384^2\, \texttt{AND} + 384 \cdot 383\, \texttt{ADD} + 384\, \texttt{MUL}} < 1.0052 
\]

For the Swin model with \( m = n = 96 \), \( l = 48 \):
\[
1 + \frac{142\, \texttt{ADD} + (3 \cdot 48 - 96)\, \texttt{MUL}}{96^2\, \texttt{AND} + 96 \cdot 95\, \texttt{ADD} + 96\, \texttt{MUL}} < 1.0207
\ 
\]

These results demonstrate that the inference computational cost of the BQQ layer is nearly equivalent to that of the conventional first-order 1-bit quantization.

\subsection{BQQ Execution Time for PTQ}\label{app:bqq_execution_time_for_ptq}

We report the quantization time required by our proposed method, BQQ, under a data-free setting. All experiments were conducted using the following environment:
\begin{itemize}
    \item Python 3.9.19
    \item PyTorch 2.6.0 with CUDA 12.4
    \item Four NVIDIA GeForce RTX 4090 GPUs
    \item AMD EPYC 7313 16-Core Processor
\end{itemize}
During quantization, we parallelized the process by assigning each matrix to a separate GPU thread, enabling concurrent quantization of multiple layers. 
Quantization time was measured using Python's \texttt{time} module.
We evaluated quantization time using a small model (DeiT-S, 22M parameters) and a large model (DeiT-B, 86M parameters).

Tab.~\ref{tab:quant_time} summarizes the computation time for pseudo 2*-bit, 3*-bit, and 4*-bit quantization using BQQ. The reported times indicate the total time required to quantize the entire model.
Note that the larger the pseudo bit width, the longer it takes because of greedy optimization for each pseudo bit index, as shown in Alg.~\ref{alg:greedy_qq}
Since processing is performed in parallel on each GPU, speedup is possible by increasing the number of GPUs, but computation time is a barrier to scaling up the applicability of PTQ to large-scale models such as large language models.

\begin{table}[h]
    \centering
        \caption{Execution time of BQQ on DeiT-S and DeiT-B.}
        \vspace{1em}
\renewcommand{\arraystretch}{1.5}
\begin{tabular}{cccrrr}
\hline
\multirow{2}{*}{\textbf{Model}} & \multirow{2}{*}{\textbf{\#layers}} & \multirow{2}{*}{\textbf{\#parameters}} & \multicolumn{3}{c}{\textbf{BQQ  Time {[}min{]}}} \\
                                &                                    &                                        & 2* bit            & 3* bit            & 4* bit            \\ \hline\hline
\textbf{DeiT-S}                 & \multicolumn{1}{r}{12}             & \multicolumn{1}{r}{22M}                & 13                & 17                & 21              \\
\textbf{DeiT-B}                 & \multicolumn{1}{r}{12}             & \multicolumn{1}{r}{86M}                & 32                & 45                & 57                \\ \hline
\end{tabular}

    \label{tab:quant_time}
\end{table}

\subsection{Effect of $N_\text{step}$ on Accuracy and Computation Time}

\begin{figure}[h]
    \centering
    \includegraphics[width=0.8\linewidth]{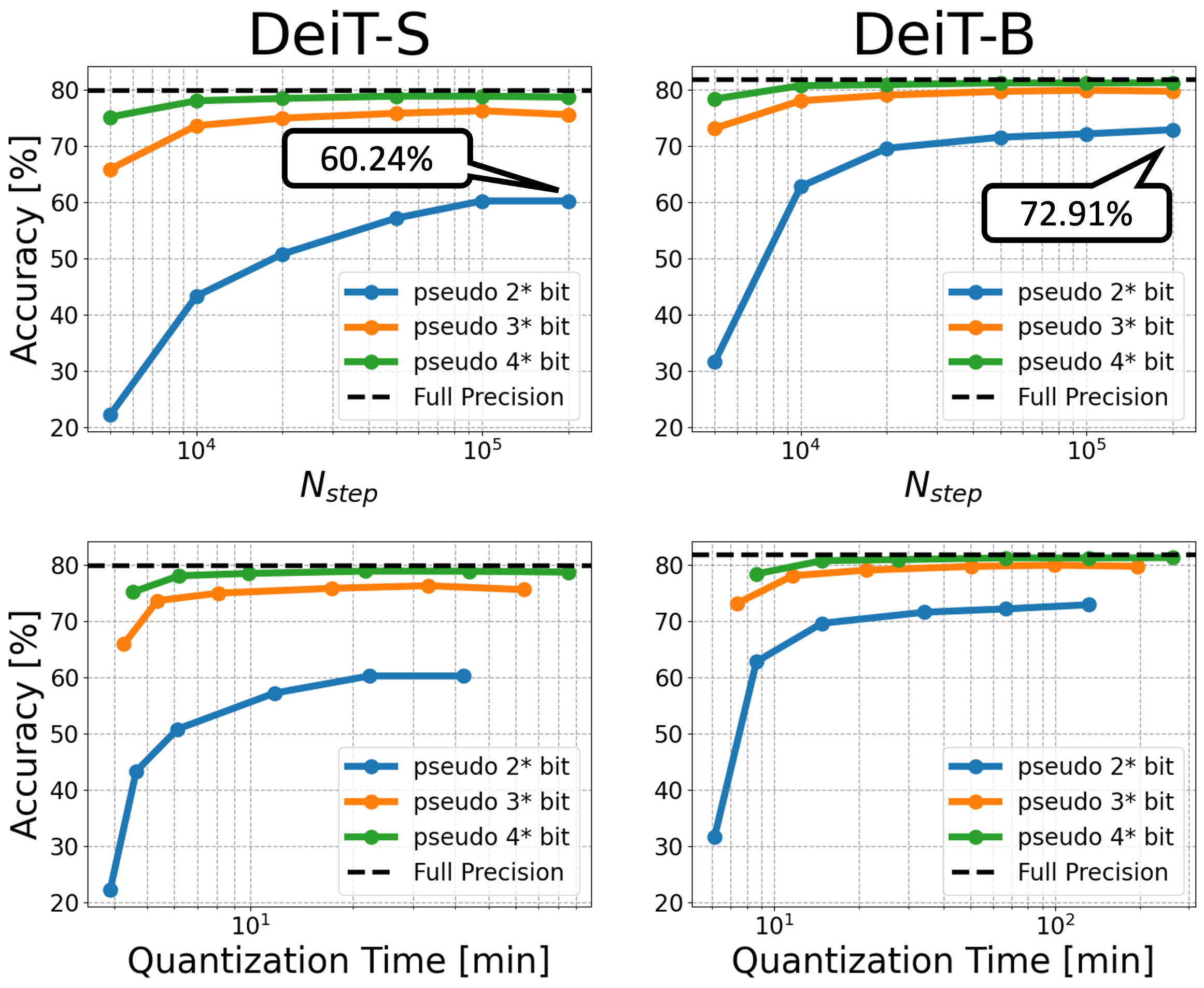}
    \caption{Effect of the number of optimization steps $N_\text{step}$ (top) and total quantization time (bottom) on final ImageNet top-1 accuracy in the data-free quantization setting.}
    \label{fig:time_scaling}
\end{figure}

We investigate the effect of the number of optimization steps $N_\text{step}$ (and the total quantization time) on the final ImageNet top-1 accuracy and computation time under the data-free quantization setting. The results on DeiT-Small (DeiT-S) and DeiT-Base (DeiT-B) models are shown in Fig.~\ref{fig:time_scaling}.
As expected, increasing $N_\text{step}$ generally results in longer quantization time but also enables more precise optimization, which tends to improve final accuracy.  
In particular, it is noteworthy that increasing $N_\text{step}$ beyond 50{,}000---the setting used in the main manuscript---yields even more accurate results. For instance, with DeiT-S, an accuracy of 60.24\% is achieved at pseudo 2*-bit precision, which is 1.99\% higher than the accuracy reported in the main manuscript. Similarly, with DeiT-B, an accuracy of 72.91\% is obtained at pseudo 2*-bit precision, representing a 0.82\% improvement.
As shown in the figure, in highly compressed settings such as pseudo 2*-bit quantization, a large $N_\text{step}$ is essential to prevent significant accuracy degradation. In contrast, for pseudo 3*-bit or 4*-bit quantization, accuracy remains relatively stable even with smaller $N_\text{step}$ values.

These observations suggest that more aggressive compression schemes are more sensitive to the quality of optimization, as even small quantization errors can have a greater impact on final accuracy. Therefore, more rigorous optimization is required in such cases.  
Notably, the time required for accuracy to reach saturation appears largely independent of the pseudo bit-width, with convergence observed in approximately 10 minutes for DeiT-S and 20 minutes for DeiT-B---durations that are not prohibitive in practical applications.

\subsection{Accuracy Under Extreme Compression}
We present additional results beyond those in the main manuscript, focusing on further compression--specifically, setting the model size to pseudo 1.5* bit. In the main experiments, the factorization was performed using binary matrices $\boldsymbol{Y}_i \in \{0,1\}^{m \times l}$ and $\boldsymbol{Z}_i \in \{0,1\}^{l \times n}$, where the intermediate dimension was set to $l = \text{round}\left(\frac{mn}{m+n}\right)$. Here, we increase the intermediate dimension to $l = \text{round}\left(1.5 \cdot \frac{mn}{m+n}\right)$ (i.e., 1.5 times the original parameter count), and fix the number of stacks (denoted as $p$ in Eq.~\eqref{eq:qq_approx}) to 1. This results in an effectively 1.5-bit quantized model, noting that all weight elements remain binary (1-bit).
Tab.~\ref{tab:1.5bit_result} reports the top-1 accuracy on ImageNet after compression. While the accuracy significantly deteriorates in the data-free setting, it is remarkable that with only a small amount of calibration data, the accuracy remains reasonably high. 
Notably, for DeiT-B, the top-1 accuracy exceeds 70\%, demonstrating that even under such severe compression, practical accuracy can be retained—an impressive outcome.

\begin{table}[h]
\centering
\caption{ImageNet top-1 accuracy of BQQ under extreme compression.}
\label{tab:1.5bit_result}
\renewcommand{\arraystretch}{1.5}
\begin{tabular}{ccccrrr}
\hline
\multirow{2}{*}{\textbf{Method}} & \multirow{2}{*}{\textbf{W bit}} & \multirow{2}{*}{\textbf{Data Free}} & \multicolumn{4}{c}{\textbf{ImageNet Top-1 Accuracy {[}\%{]}}}                                                                               \\
                                 &                                 &                                     & \textbf{DeiT-S}           & \multicolumn{1}{c}{\textbf{DeiT-B}} & \multicolumn{1}{c}{\textbf{Swin-T}} & \multicolumn{1}{c}{\textbf{Swin-S}} \\ \hline\hline
\textbf{c-BQQ}                   & 1.5*                            & $\times$                                   & \multicolumn{1}{r}{53.67} & 71.36                               & 61.25                               & 69.89                               \\
\textbf{BQQ}                     & 1.5*                            & $\checkmark$                                   & \multicolumn{1}{r}{7.41}  & 35.21                               & 10.82                               & 20.06                               \\ \hline
\end{tabular}
\end{table}

\subsection{Evaluation on Language Models}
\begin{table*}[]
\centering
\caption{WikiText-2 perplexity and downstream task accuracy.}
\label{tab:evaluation_on_language_models}
\fontsize{7.5pt}{7.5pt}\selectfont
\renewcommand{\arraystretch}{1.5}
\begin{tabular}{cccrrrrrrrr}
\hline
Model                                                                                     & Method     & Bit & \multicolumn{1}{c}{PPL} & \multicolumn{1}{c}{ArcE} & \multicolumn{1}{c}{ArcC} & \multicolumn{1}{c}{BoolQ} & \multicolumn{1}{c}{HellaS} & \multicolumn{1}{c}{PiQA} & \multicolumn{1}{c}{WinoG} & \multicolumn{1}{c}{Avg.} \\ \hline\hline
\multirow{10}{*}{Qwen2.5-0.5B}                                                            & BQQ   & 2   & 17106.9                 & 28.3                      & 20.3                      & 37.8                      & 26.3                       & 54.0                     & 50.5                      & 36.2                     \\
                                                                                          & t-BQQ & 2   & \textbf{49.5}           & \textbf{37.1}             & \textbf{24.2}             & 39.7                      & \textbf{29.9}              & \textbf{56.4}            & \textbf{52.1}             & \textbf{39.9}            \\
                                                                                          & GPTQ  & 2   & 4392.5                  & 25.3                      & 21.3                      & \textbf{43.2}             & 25.7                       & 52.7                     & 49.0                      & 36.2                     \\ \cline{2-11} 
                                                                                          & BQQ   & 3   & 1808.8                  & 31.5                      & 19.1                      & 38.0                      & 26.5                       & 54.4                     & 50.5                      & 36.7                     \\
                                                                                          & t-BQQ & 3   & \textbf{19.4}           & 45.2                      & \textbf{27.2}             & 43.0                      & \textbf{39.7}              & \textbf{62.1}            & \textbf{56.1}             & 45.6                     \\
                                                                                          & GPTQ  & 3   & 21.8                    & \textbf{45.8}             & 21.0                      & \textbf{56.5}             & 34.0                       & \textbf{63.6}            & 56.0                      & \textbf{46.1}            \\ \cline{2-11} 
                                                                                          & BQQ   & 4   & 71.7                    & 38.8                      & 21.6                      & 43.9                      & 30.5                       & 61.2                     & 51.2                      & 41.2                     \\
                                                                                          & t-BQQ & 4   & 14.7                    & 51.4                      & \textbf{30.5}             & \textbf{52.7}             & \textbf{45.6}              & \textbf{66.6}            & 54.3                      & 50.2                     \\
                                                                                          & GPTQ  & 4   & \textbf{14.4}           & \textbf{62.9}             & 27.9                      & \textbf{57.4}             & 38.9                       & \textbf{68.2}            & \textbf{56.2}             & \textbf{51.9}            \\ \cline{2-11} 
                                                                                          & baseline  & 16  & 13.1                    & 64.6                      & 29.5                      & 58.8                      & 40.6                       & 70.2                     & 56.4                      & 53.4                     \\ \hline
\multirow{10}{*}{Qwen2.5-1.5B}                                                            & BQQ   & 2   & 688.3                   & 33.8                      & 19.1                      & 57.7                      & 27.2                       & 56.6                     & 51.9                      & 41.0                     \\
                                                                                          & t-BQQ & 2   & \textbf{23.6}           & \textbf{45.1}             & \textbf{28.5}             & \textbf{62.0}             & \textbf{40.2}              & \textbf{61.2}            & \textbf{53.8}             & \textbf{48.5}            \\
                                                                                          & GPTQ  & 2   & 922.4                   & 26.0                      & 21.1                      & 45.0                      & 26.0                       & 51.6                     & 52.2                      & 37.0                     \\ \cline{2-11} 
                                                                                          & BQQ   & 3   & 14.6                    & \textbf{63.6}             & 32.0                      & \textbf{63.9}             & 41.7                       & 71.1                     & 57.9                      & 55.0                     \\
                                                                                          & t-BQQ & 3   & \textbf{11.8}           & 61.8                      & \textbf{38.3}             & 60.6                      & \textbf{57.0}              & \textbf{72.0}            & \textbf{59.0}             & \textbf{58.1}            \\
                                                                                          & GPTQ  & 3   & 12.1                    & 59.5                      & 30.0                      & \textbf{60.7}             & 43.6                       & 70.0                     & 57.1                      & 53.5                     \\ \cline{2-11} 
                                                                                          & BQQ   & 4   & 10.5                    & 72.4                      & 38.5                      & 69.1                      & 47.7                       & 74.8                     & 60.7                      & 60.5                     \\
                                                                                          & t-BQQ & 4   & 9.9                     & 73.0                      & \textbf{45.1}             & 65.5                      & \textbf{62.9}              & 73.8                     & 61.3                      & \textbf{63.6}            \\
                                                                                          & GPTQ  & 4   & \textbf{9.7}            & \textbf{74.5}             & 40.0                      & \textbf{70.8}             & 49.3                       & \textbf{75.2}            & \textbf{63.1}             & 62.1                     \\ \cline{2-11} 
                                                                                          & baseline  & 16  & 9.3                     & 71.5                      & 45.1                      & 73.0                      & 67.8                       & 76.1                     & 63.4                      & 66.1                     \\ \hline
\multirow{10}{*}{\begin{tabular}[c]{@{}c@{}}Deepseek-R1-Distill\\ Qwen-1.5B\end{tabular}} & BQQ   & 2   & 921.2                   & 29.8                      & 18.8                      & 47.9                      & 26.5                       & 53.6                     & 49.3                      & 37.6                     \\
                                                                                          & t-BQQ & 2   & \textbf{46.1}           & \textbf{41.1}             & \textbf{22.7}             & \textbf{52.5}             & \textbf{31.0}              & \textbf{58.8}            & \textbf{52.6}             & \textbf{43.1}            \\
                                                                                          & GPTQ  & 2   & 872.4                   & 26.6                      & 20.2                      & 47.6                      & 25.7                       & 52.9                     & 49.1                      & 37.0                     \\ \cline{2-11} 
                                                                                          & BQQ   & 3   & 59.2                    & \textbf{56.1}             & 30.6                      & 51.2                      & 33.8                       & \textbf{63.2}            & 52.9                      & 48.0                     \\
                                                                                          & t-BQQ & 3   & \textbf{28.5}           & 50.7                      & \textbf{32.9}             & \textbf{59.3}             & \textbf{41.2}              & 61.7                     & \textbf{55.7}             & \textbf{50.3}            \\
                                                                                          & GPTQ  & 3   & 59.6                    & 52.3                      & 25.6                      & 54.0                      & 32.8                       & 62.4                     & 51.9                      & 46.5                     \\ \cline{2-11} 
                                                                                          & BQQ   & 4   & \textbf{39.4}           & \textbf{59.3}             & 31.8                      & 66.5                      & 36.0                       & 64.8                     & \textbf{57.5}             & \textbf{52.7}            \\
                                                                                          & t-BQQ & 4   & \textbf{21.7}           & 54.1                      & 31.9                      & 65.2                      & \textbf{43.9}              & 63.8                     & 56.0                      & 52.5                     \\
                                                                                          & GPTQ  & 4   & 43.4                    & 58.5                      & \textbf{32.2}             & \textbf{66.9}             & 36.0                       & \textbf{65.9}            & 56.3                      & 52.6                     \\ \cline{2-11} 
                                                                                          & baseline  & 16  & 40.4                    & 56.1                      & 34.6                      & 68.6                      & 44.8                       & 65.8                     & 55.6                      & 54.2                     \\ \hline
\end{tabular}
\end{table*}

Here, we evaluate our quantization methods on several compact language models that are well-suited for edge deployment: Qwen2.5-0.5B, Qwen2.5-1.5B, and DeepSeek-R1-Distill-Qwen1.5B.
We compare three approaches:
(1) BQQ, which performs data-free quantization;
(2) tuned-BQQ (t-BQQ), which first applies BQQ and then fine-tunes only the continuous parameters; and
(3) GPTQ~\cite{frantar2023gptq}, a standard post-training quantization (PTQ) method.
For both t-BQQ and GPTQ, the calibration data consist of the full training split of WikiText-2.

Tab.~\ref{tab:evaluation_on_language_models} reports the perplexity on WikiText-2~\cite{Merity2016_wikitext2} and the accuracy on six downstream tasks: PIQA~\cite{Bisk2019PIQARA}, Winogrande (WinoG)~\cite{Sakaguchi2019_winog}, ARC-Easy (ArcE) and ARC-Challenge (ArcC)~\cite{Clark2018Arc}, HellaSwag (HellaS)~\cite{Zellers2019HellaSwagCA}, and BoolQ~\cite{Clark2019BoolQET}.
The table also includes the average accuracy across all tasks.
BQQ achieves higher average accuracy than GPTQ on the 1.5B models, particularly under low-bit settings (2-bit and 3-bit).
Notably, even without any calibration data, BQQ attains comparable or superior performance to GPTQ.
In contrast, for the smaller 0.5B model, BQQ struggles to maintain accuracy.
We attribute this to the model-dependent discrepancy between the quantization distributions that minimize activation error and those that minimize weight error.
When this discrepancy becomes large, the performance of BQQ tends to degrade.

Although the performance of neural network quantization is not always guaranteed to be superior, we emphasize that the current work proposes BQQ as a general binary quantization framework rather than a dedicated quantization technique for neural networks. In our implementation, BQQ quantizes weights by minimizing reconstruction error in the weight space, without relying on output-based error signals that are commonly used in many neural network quantization methods. This implies that BQQ does not yet exploit task-specific loss functions or activation statistics. We therefore believe that adapting BQQ to incorporate such feedback—especially minimizing the downstream output error—could lead to significant further improvements in model performance, and this represents a highly promising avenue for future research.

\subsection{Quantization Error vs. Binary Matrix Stack-to-Intermediate Dimension Ratio}\label{app:error-stack-width}

In the main manuscript, we fixed the intermediate dimension as 
$l = \mathrm{round}\!\left(\frac{mn}{m+n}\right)$ 
for all experiments. 
However, this setting is not necessarily optimal.
The same compression ratio can also be achieved by varying 
the number of stacked binary matrices and the intermediate dimension. 
In this subsection, we investigate how the quantization error (MSE) changes 
with respect to the number of stacks $p$ and the normalized intermediate dimension 
$l_\text{scale}$, where the actual intermediate dimension is defined as 
$l = \mathrm{round}\!\left(l_\text{scale}\frac{mn}{m+n}\right)$. 
We conducted experiments by sweeping over $p$ and $l_\text{scale}$ 
and measuring the resulting MSE. 

Tab.~\ref{tab:stack-wide-ratio} summarizes the experimental results. 
These results show that the optimal balance between the number of stacks 
and the intermediate dimension varies across datasets, 
indicating that fixing $l = \mathrm{round}\!\left(\frac{mn}{m+n}\right)$ 
is not always optimal. 
Therefore, adaptively determining this ratio can potentially lead to 
more efficient compression.
\begin{table*}[h]
\fontsize{6.5pt}{6.5pt}\selectfont
\centering
\caption{Quantization error (MSE) with varying the number of stacked binary matrices $p$ 
and the intermediate dimension scaling $l_\text{scale}$.}
\label{tab:stack-wide-ratio}
\renewcommand{\arraystretch}{1.5}
\begin{tabular}{|>{\columncolor[gray]{0.8}}r>{\columncolor[gray]{0.8}}r|rr|rr|rr|rr|rr|}
\hline
\rowcolor[gray]{0.8}
\multicolumn{1}{|c}{}             & \multicolumn{1}{c|}{}        & \multicolumn{2}{c|}{\textbf{Random}}                                & \multicolumn{2}{c|}{\textbf{DeiT}}                                    & \multicolumn{2}{c|}{\textbf{Distance}}                                & \multicolumn{2}{c|}{\textbf{SIFT}}                                    & \multicolumn{2}{c|}{\textbf{ImageNet}}                                \\
\rowcolor[gray]{0.8}
\multicolumn{1}{|c}{\textbf{\#stacks ($p$)}} & \multicolumn{1}{c|}{\textbf{$l_\text{scale}$}} & \multicolumn{1}{c}{\textbf{Size {[}KB{]}}} & \multicolumn{1}{c|}{\textbf{MSE}} & \multicolumn{1}{c}{\textbf{Size {[}KB{]}}} & \multicolumn{1}{c|}{\textbf{MSE}} & \multicolumn{1}{c}{\textbf{Size {[}KB{]}}} & \multicolumn{1}{c|}{\textbf{MSE}} & \multicolumn{1}{c}{\textbf{Size {[}KB{]}}} & \multicolumn{1}{c|}{\textbf{MSE}} & \multicolumn{1}{c}{\textbf{Size {[}KB{]}}} & \multicolumn{1}{c|}{\textbf{MSE}} \\ \hline
1                                 & 1                            & 2.1                               & \textbf{324.3}                    & 18.4                              & \textbf{298.1}                    & 1.3                               & 14.6                     & 2.1                               & \textbf{97.8}                     & 6.3                               & 42.7                     \\
2                                 & 0.5                          & 2.1                               & 329.9                    & 18.5                              & 309.4                    & 1.3                               & \textbf{9.2}                      & 2.1                               & 108.1                    & 6.3                               & \textbf{29.7}                     \\
4                                 & 0.25                         & 2.1                               & 337.1                    & 18.5                              & 317.3                    & 1.3                               & 12.2                     & 2.1                               & 110.8                    & 6.3                               & 49.6                     \\ \hline
1                                 & 2                            & 4.1                               & 106.4                    & 36.9                              & 101.9                    & 2.5                               & 7.7                      & 4.1                               & 46.3                     & 12.6                              & 20.5                     \\
2                                 & 1                            & 4.1                               & \textbf{105.3}                    & 36.9                              & \textbf{95.8}                     & 2.5                               & \textbf{2.3}                      & 4.1                               & \textbf{30.0}                     & 12.6                              & 10.9                     \\
4                                 & 0.5                          & 4.1                               & 108.4                    & 36.9                              & 100.8                    & 2.6                               & 2.5                      & 4.1                               & 34.4                     & 12.6                              & \textbf{8.9}                      \\
8                                 & 0.25                         & 4.2                               & 114.2                    & 37.0                              & 105.6                    & 2.5                               & 3.5                      & 4.2                               & 35.9                     & 12.6                              & 14.1                     \\ \hline
2                                 & 1.5                          & 6.2                               & \textbf{33.2}                     & 55.3                              & \textbf{30.6}                     & 3.8                               & 1.3                      & 6.2                               & 10.9                     & 18.8                              & 4.2                      \\
3                                 & 1                            & 6.2                               & 34.4                     & 55.3                              & 31.0                     & 3.8                               & \textbf{0.7}                      & 6.2                               & \textbf{9.7}                      & 18.9                              & 3.5                      \\
6                                 & 0.5                          & 6.2                               & 36.0                     & 55.4                              & 33.2                     & 3.8                               & 0.9                      & 6.2                               & 11.5                     & 18.9                              & \textbf{2.9}                      \\
12                                & 0.25                         & 6.3                               & 38.4                     & 55.4                              & 35.7                     & 3.7                               & 1.2                      & 6.3                               & 12.2                     & 19.0                              & 4.7                      \\ \hline
2                                 & 2                            & 8.2                               & 11.5                     & 73.8                              & 10.9                     & 5.0                               & 0.6                      & 8.2                               & 4.9                      & 25.1                              & 2.1                      \\
4                                 & 1                            & 8.2                               & \textbf{11.2}                     & 73.8                              & \textbf{10.0}                     & 5.1                               & \textbf{0.2}                      & 8.2                               & \textbf{3.2}                      & 25.1                              & 1.1                      \\
8                                 & 0.5                          & 8.3                               & 11.9                     & 73.8                              & 10.9                     & 5.1                               & 0.3                      & 8.3                               & 3.8                      & 25.2                              & \textbf{1.0}                      \\
16                                & 0.25                         & 8.4                               & 13.0                     & 73.9                              & 12.0                     & 5.0                               & 0.4                      & 8.4                               & 4.1                      & 25.3                              & 1.6                      \\ \hline
\end{tabular}
\end{table*}

\subsection{Theoretical Upper Bound of BQQ}
\label{app:approx_bound}

In this subsection, we provide a theoretical analysis of the approximation error inherent in the BQQ formulation.
To derive a concrete upper bound, we consider a particular case of Eq.~\eqref{eq:before_BQQ} by setting $\beta_i = -0.5\alpha_i$, $\delta_i = 1$, and $\gamma_i = -0.5$.
Although this specific case does not yield a closed-form optimal solution, an approximate one can be obtained by aligning the binary components with the sign patterns of the singular vectors obtained from SVD, combined with appropriate scaling.
This leads to a theoretical upper bound on the square root of the subproblem error, denoted by $\sqrt{L_{\text{sub}}^{(i)}}$ (Eq.~\eqref{eq:qq_sub}).
The overall error of BQQ is then obtained by summing over all stack indices $i$:
$\text{BQQ total error} = \sum_i L_{\text{sub}}^{(i)}$.
Based on the Eckart--Young--Mirsky theorem and the triangle inequality, we obtain the following bound:
\small
\begin{align}
\text{min }\sqrt{L_{\text{sub}}^{(i)}} &\le \text{min } \left\| \boldsymbol{W}^{(i)} - \left[ \alpha_i( \boldsymbol{Y}_i - 0.5\cdot\boldsymbol{1}_Y)( \boldsymbol{Z}_i - 0.5\cdot\boldsymbol{1}_Z)\right] \right\|\\
&= \text{min } \left\| \boldsymbol{W}^{(i)} - \boldsymbol{W}_{\text{svd}}^{(i)} + \boldsymbol{W}_{\text{svd}}^{(i)} - \left[ \alpha_i( \boldsymbol{Y}_i - 0.5\cdot\boldsymbol{1}_Y)( \boldsymbol{Z}_i - 0.5\cdot\boldsymbol{1}_Z)\right] \right\| \\
&\le \text{min } \left\| \boldsymbol{W}^{(i)} - \boldsymbol{W}_{\text{svd}}^{(i)} \right\|
  + \left\| \boldsymbol{W}_{\text{svd}}^{(i)} - \left[ \alpha_i( \boldsymbol{Y}_i - 0.5\cdot\boldsymbol{1}_Y)( \boldsymbol{Z}_i - 0.5\cdot\boldsymbol{1}_Z)\right] \right\| \\
&\le \text{min } \sqrt{\sum_{j=l+1}^{\text{min}(m,n)}\sigma_j^2}
  + \left\| \boldsymbol{W}_{\text{svd}}^{(i)} - \alpha_i \cdot \text{sgn}\left(\boldsymbol{U}_{\text{svd}}^{(i)}\right) \text{sgn}\left(\boldsymbol{V}_{\text{svd}}^{(i)}\right) \right\| \\
&= \sqrt{\sum_{j=l+1}^{\text{min}(m,n)}\sigma_j^2}
  + \left\| \boldsymbol{W}_{\text{svd}}^{(i)} -
    \frac{\left\langle \boldsymbol{W}_{\text{svd}}^{(i)},
    \text{sgn}\left(\boldsymbol{U}_{\text{svd}}^{(i)}\right)
    \text{sgn}\left(\boldsymbol{V}_{\text{svd}}^{(i)}\right)\right\rangle}
    {\left\|\text{sgn}\left(\boldsymbol{U}_{\text{svd}}^{(i)}\right)
    \text{sgn}\left(\boldsymbol{V}_{\text{svd}}^{(i)}\right)\right\|^2}
    \cdot \text{sgn}\left(\boldsymbol{U}_{\text{svd}}^{(i)}\right)
    \text{sgn}\left(\boldsymbol{V}_{\text{svd}}^{(i)}\right)
  \right\| ,
\label{eq:approx_bound_bqq}
\end{align}
\normalsize
where $\sigma_j$ and $\boldsymbol{W}_{\text{svd}}^{(i)}$ denote the singular values and the reconstructed matrix after applying an $l$-rank approximation to the SVD of $\boldsymbol{W}^{(i)}$, respectively.
Furthermore, $\boldsymbol{U}_{\text{svd}}^{(i)}$ and $\boldsymbol{V}_{\text{svd}}^{(i)}$ represent the left and right singular vectors obtained from the SVD of $\boldsymbol{W}^{(i)}$, where $\boldsymbol{U}_{\text{svd}}^{(i)}$ already incorporates the top $l$ singular values.

This result indicates that the approximation error of the BQQ formulation is closely related to the magnitude of the discarded singular values.
Consequently, BQQ achieves higher representational fidelity for matrices with rapidly decaying singular spectra, while its approximation quality degrades when the low-rank truncation leaves substantial residual energy.
This theoretical property can also be observed in the trends shown in Fig.~\ref{fig:mse_matrix}.


\end{document}